\documentclass[lettersize,journal]{IEEEtran}
\usepackage[T1]{fontenc}
\usepackage[table]{xcolor} 

\usepackage{amsmath,amsfonts}
\usepackage{algorithmic}
\usepackage{algorithm}
\usepackage{array}
\usepackage[caption=false,font=normalsize,labelfont=sf,textfont=sf]{subfig}
\usepackage{textcomp}
\usepackage{stfloats}
\usepackage{url}
\usepackage{verbatim}
\usepackage{graphicx}
\usepackage{cite}
\usepackage{xspace}
\usepackage{xcolor}
\usepackage{bbding}
\usepackage{lineno}

\usepackage{booktabs}
\usepackage{multirow}
\usepackage{arydshln}
\usepackage{enumitem}

\usepackage{color}
\usepackage{colortbl}

\newcommand{\cq}{\textcolor{black}}

\usepackage[colorlinks,linkcolor=blue]{hyperref}

\usepackage{float}

\begin{document}

\title{Precise GPS-Denied UAV Self-Positioning via Context-Enhanced Cross-View Geo-Localization}

\author{Yuanze Xu, Ming Dai, Wenxiao Cai and Wankou Yang,~\IEEEmembership{Member,~IEEE,}
  \thanks{This work was supported in part by the National Natural Science Foundation of China (NSFC) under Grant 62276061. (\textit{Corresponding author: Wankou Yang}.)}
  \thanks{Yuanze Xu, Ming Dai and Wankou Yang are with the School of Automation and the Key Laboratory of Measurement and Control of Complex Systems of Engineering, Southeast University, Nanjing, Jiangsu Province, 210096, China (e-mail: 220232101@seu.edu.cn; mingdai@seu.edu.cn; wkyang@seu.edu.cn). Wankou Yang is also with the Advanced Ocean Institute of Southeast University, Nantong, Jiangsu Province, 226019, China.}
  \thanks{Wenxiao Cai is with Stanford University, Stanford, CA 94305, USA. (email: wxcai@stanford.edu).}
}


\markboth{Journal of \LaTeX\ Class Files,~Vol.~14, No.~8, August~2021}%
{Shell \MakeLowercase{\textit{et al.}}: A Sample Article Using IEEEtran.cls for IEEE Journals}


\maketitle

\begin{abstract}
  Image retrieval has been employed as a robust complementary technique to address the challenge of Unmanned Aerial Vehicles (UAVs) self-positioning.
  However, most existing methods primarily focus on localizing objects captured by UAVs through complex part-based representations, often overlooking the unique challenges associated with UAV self-positioning, such as fine-grained spatial discrimination requirements and dynamic scene variations.
  To address the above issues, we propose the \underline{C}ontext-\underline{E}nhanced method for precise \underline{U}AV \underline{S}elf-\underline{P}ositioning (CEUSP), specifically designed for UAV self-positioning tasks.
  CEUSP integrates a Dynamic Sampling Strategy (DSS) to efficiently select optimal negative samples, while the Rubik's Cube Attention (RCA) module, combined with the Context-Aware Channel Integration (CACI) module, enhances feature representation and discrimination by exploiting interdimensional interactions, inspired by the rotational mechanics of a Rubik's Cube.
  Extensive experimental validate the effectiveness of the proposed method, demonstrating notable improvements in feature representation and UAV self-positioning accuracy within complex urban environments.
  Our approach achieves state-of-the-art performance on the DenseUAV dataset, which is specifically designed for dense urban contexts, and also delivers competitive results on the widely recognized University-1652 benchmark.
\end{abstract}


\section{Introduction}
\IEEEPARstart{U}{NMANNED} Aerial Vehicles (UAVs) have received considerable attention in recent years owing to their superior data acquisition capabilities.
As advanced visual platforms~\cite{7508986, pliego2021quaternion, 10256085}, UAVs are capable of capturing extensive imagery from diverse angles while effectively mitigating the impact of occlusions.
This ability has led to widespread applications, including object detection~\cite{10158513, 10119181, 9119818, dai2024simvg}, precise delivery systems~\cite{6668877, 10620851, 9933782}, and autonomous driving technologies~\cite{9700861, 10194407, 9919263}.
However, effective self-positioning is essential for UAVs to operate successfully across these domains.
Currently, most UAVs rely on the Global Positioning System (GPS) technology for navigation, which requires a stable communication environment.
Disruptions or obstructions to GPS signals can significantly degrade positioning accuracy, limiting operational effectiveness in certain contexts.
In response to these challenges, image-based cross-view geo-location techniques serve as viable alternatives~\cite{10582419}, enhancing the reliability of UAV self-positioning in GPS-denied environments.

\begin{figure}[!t]
  \centering
  \includegraphics[width=0.95\linewidth]{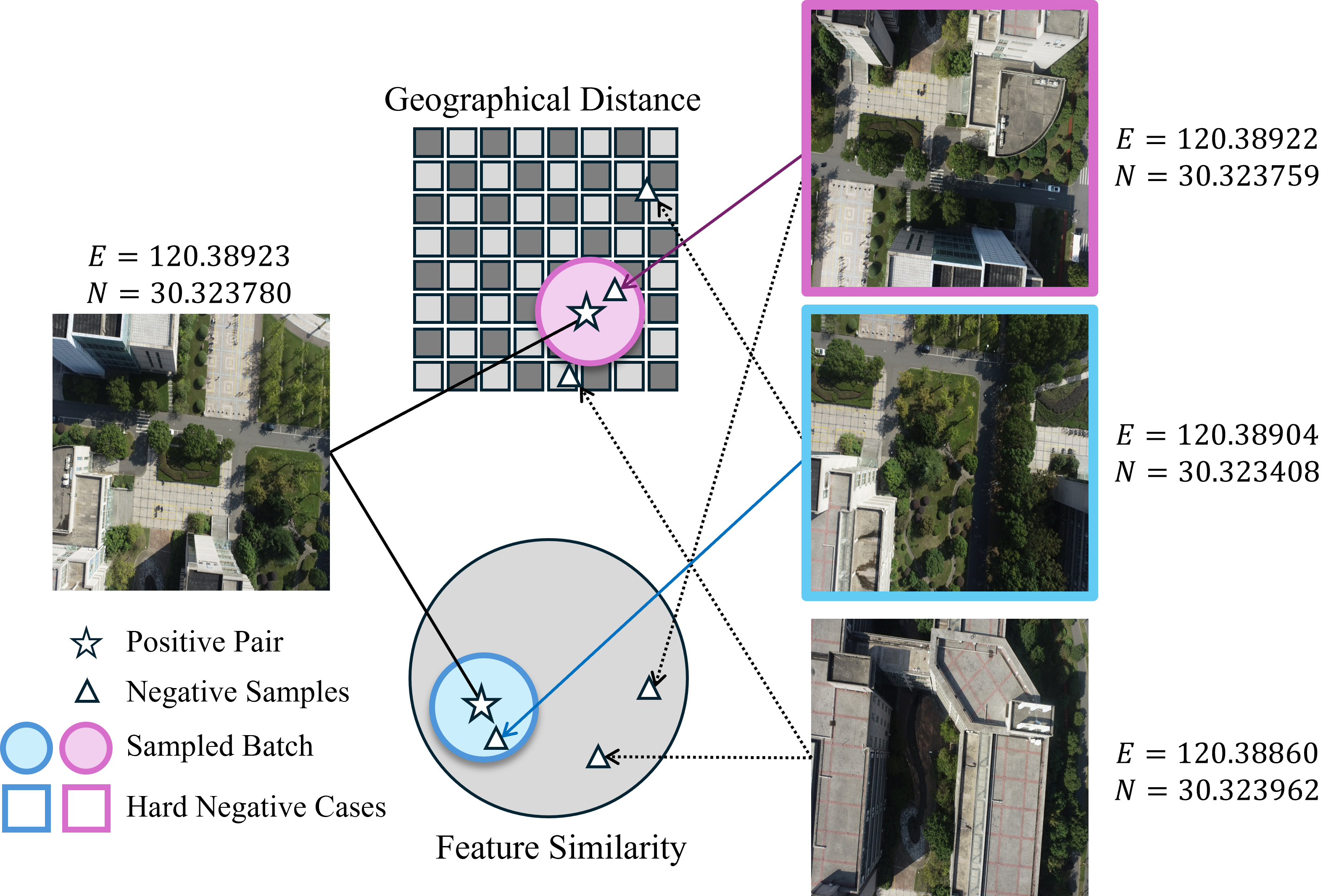}
  \caption{The Dynamic Sampling Strategy. Purple samples are selected from images exhibiting geographic proximity, while blue samples are derived from feature similarities identified between drone and satellite imagery. Within each sampling batch, the ratio of these two sample types is adjusted progressively as the training process advances.}
  \label{sampler}
  \vspace{-15pt}
\end{figure}

The cross-view geographic localization technology achieves precise positioning by correlating images obtained from diverse perspectives.
However, the significant visual discrepancies between these viewpoints pose substantial challenges for cross-view image matching, creating a notable domain gap between ground and satellite imagery.
Drone-captured images emphasize intricate ground-level details, such as building facades, pedestrians, and vehicles, while satellite imagery focuses on large-scale geospatial features like rooftops, tree canopies, and road networks.
Despite these considerable appearance disparities, both types of imagery possess the potential to convey rich semantic information.
Previous research has primarily concentrated on matching from ground-to-aerial~\cite{10177194, 9913952, 10601183} and drone-to-satellite~\cite{wang2021each, 10387514, 9583266} perspectives, primarily addressing the localization of objects captured by drones.
However, challenges associated with UAV self-positioning, such as low-altitude dense sampling and variations in urban surface appearance~\cite{10376356}, remain underexplored.
Methods employing part-based representation learning~\cite{wang2021each, chen2024sdpl} show promise on sparse datasets like University-1652~\cite{zheng2020university}, but struggle in densely sampled urban contexts, where they disrupt image structure and incur semantic information loss.
While Shen \textit{et al.}~\cite{10185134} enhance feature discrimination with a multi-classifier structure, they face limitations in densely sampled scenes due to limited cross-dimensional interactions.

Traditional transformer- and part-based models~\cite{wang2021each, 9779991} often enhance representation by segmenting the feature map, focusing on local regions without capturing multi-dimensional feature interactions, leading to suboptimal performance in UAV self-positioning tasks.
Our method, Context-Enhanced method for precise UAV Self-Positioning (CEUSP), transcends this limitation by rethinking feature extraction through the lens of "multi-dimensional interaction," moving beyond conventional "feature cutting" approaches.
The Rubik's Cube Attention (RCA) module achieves this by rotating the feature map to integrate information from multiple directions, thereby enhancing feature extraction at higher levels of abstraction.
Additionally, CEUSP incorporates a Dynamic Sampling Strategy, as shown in Fig.~\ref{sampler}, which prioritizes challenging negative samples and balances geographic relevance and feature diversity within each training batch.
This adaptive strategy ensures more efficient learning as training progresses, improving the model's performance on dense datasets.
By combining these advancements, CEUSP demonstrates broad applicability, extending its utility to other cross-view geo-localization tasks beyond UAV self-positioning.

In summary, the primary contributions of this paper are as follows:
\begin{itemize}
  \item
        Augmented by the Rubik's Cube Attention module, our ConvNeXt-T-based framework enhances feature representation by enabling efficient interactions across spatial and channel dimensions for UAV self-positioning and cross-view geo-localization tasks.
  \item
        To significantly enhance model robustness and improve feature matching accuracy, we propose a dynamic sampling strategy that adaptively refines sample selection by balancing the geographic relevance of samples with feature diversity within each training batch throughout the training process.
  \item
        Our proposed model achieves state-of-the-art performance in UAV self-positioning tasks on the DenseUAV dataset and demonstrates competitive results on the University-1652 dataset, thereby highlighting its robustness across various cross-view geo-localization challenges.
\end{itemize}

The remainder of this paper is organized as follows. In Section~\ref{Related Works}, we provide a review of the relevant literature. Section~\ref{Proposed Method} presents a detailed description of the proposed framework. Subsequently, Section~\ref{Experimental Results} discusses the experimental findings and offers a comprehensive analysis based on the DenseUAV and University-1652 datasets. Finally, Section~\ref{Conclusion} concludes with a summary of our methods.

\section{Related Works}\label{Related Works}
In this section, we review the most relevant research in two key areas: traditional cross-view geo-localization and drone-to-satellite matching in dense environments.

\subsection{Traditional Cross-View Geo-Localization}
Cross-view geo-localization has gained significant attention due to its wide range of applications~\cite{cheng2023ai, shadiev2023systematic, molina2023review, bakirci2024enhancing}.
Research has predominantly focused on matching ground-level and satellite images, as well as drone-to-satellite perspectives, with particular emphasis on accurately localizing objects within the drone's field of view.

Early cross-view geo-localization methods relied on hand-crafted features, treating the task as a retrieval problem by comparing aerial and ground images using metrics like Euclidean distance or cosine similarity. The advent of convolutional neural networks (CNNs)~\cite{krizhevsky2012imagenet, lecun1998gradient} brought significant progress, especially through Siamese networks~\cite{bromley1993signature, hadsell2006dimensionality}, enabling joint processing of aerial and ground images.
However, challenges due to visual domain discrepancies between ground-level and aerial imagery persisted, leading to degraded performance~\cite{hu2018cvm}.
Vision Transformer (ViT)~\cite{dosovitskiy2020image} have shown significant promise, as demonstrated by Dai \textit{et al.}~\cite{dai2021transformer}, who achieved notable improvements over CNN-based methods. Building on this, Zhu \textit{et al.}~\cite{zhu2022transgeo} proposed TransGeo, incorporating Transformer encoders for both street-level and bird's-eye view (BEV) images with a two-stage training strategy. Zhang \textit{et al.}~\cite{zhang2023cross} further enhanced feature granularity with GeoDTR, employing geometric layout extractors to capture spatial configurations. Other works, such as Deuser \textit{et al.}~\cite{deuser2023sample4geo} and Shi \textit{et al.}~\cite{shi2022cvlnet}, explored hard negative sampling strategies and domain adaptation via transfer matrices to bridge the visual domain gap between BEV and ground images. Ye \textit{et al.}~\cite{ye2024sg} and Shen \textit{et al.}~~\cite{10185134} introduced collaborative networks and multi-classifier structures, respectively, to enhance cross-view matching accuracy.
In parallel, part-based learning has gained traction. Wang \textit{et al.}~\cite{wang2021each} developed the local pattern network to maintain contextual integrity while extracting global features, while Dai \textit{et al.}~\cite{dai2021transformer} and Liu \textit{et al.}~\cite{liu2024adaptive} introduced segmentation and adaptive semantic aggregation techniques to enhance robustness and feature retention. Chen \textit{et al.}~\cite{chen2024sdpl} proposed the SDPL framework, which partitions features into diverse shapes to capture fine-grained details.

Rather than relying on part-based learning methods, our model integrates the Rubik's Cube Attention (RCA) module with the Context-Aware Channel Integration (CACI) module to achieve comprehensive global semantic extraction.
This dual-attention mechanism enhances the model's ability to interpret complex semantic structures, leading to improved drone self-positioning and overall cross-view matching performance.

\subsection{Drone-to-Satellite Matching in Dense Environments}
The rapid development of drone technology has spurred significant research into drone-based cross-view geo-localization.
A key contribution was made by Zheng \textit{et al.}~\cite{zheng2020university}, who introduced the University-1652 dataset, formalizing cross-view geo-localization for drone applications. Building on this, Zhuang \textit{et al.}~\cite{zhuang2021faster} proposed a multiscale block attention mechanism, addressing challenges like scale variations and target misalignment. Similarly, Zhu \textit{et al.}~~\cite{zhu2023sues} introduced the SUES-200 dataset to enhance model robustness in handling altitude variations in dynamic environments.
Focusing on dense urban scenarios, Dai \textit{et al.}~\cite{10376356} developed the DenseUAV dataset, tailored to low-altitude, GPS-denied environments, and further proposed a drone referring localization method~\cite{dai2022finding} to improve spatial feature interaction between UAV and satellite imagery. Wang \textit{et al.}~\cite{wang2023wamf} addressed feature misalignment with a weight-adaptive multi-feature fusion network, while Chen \textit{et al.}~\cite{chen2024fpi} introduced a coarse-to-fine one-stream feature interaction network, reducing computational overhead while enhancing performance through early-stage feature interaction.

DenseUAV poses unique challenges due to dense sampling and significant spatial overlap between adjacent frames, demanding sophisticated feature extraction techniques.
Traditional methods often struggle with UAV self-positioning in such dense environments.
Our approach further incorporates dynamic sampling strategies, which balance the dataset and substantially enhance performance in both UAV self-positioning and traditional cross-view geo-localization tasks, particularly in complex and dynamic environments.

\section{Proposed Method}\label{Proposed Method}
\begin{figure*}[!t]
  \centering
  \includegraphics[width=1.0\linewidth]{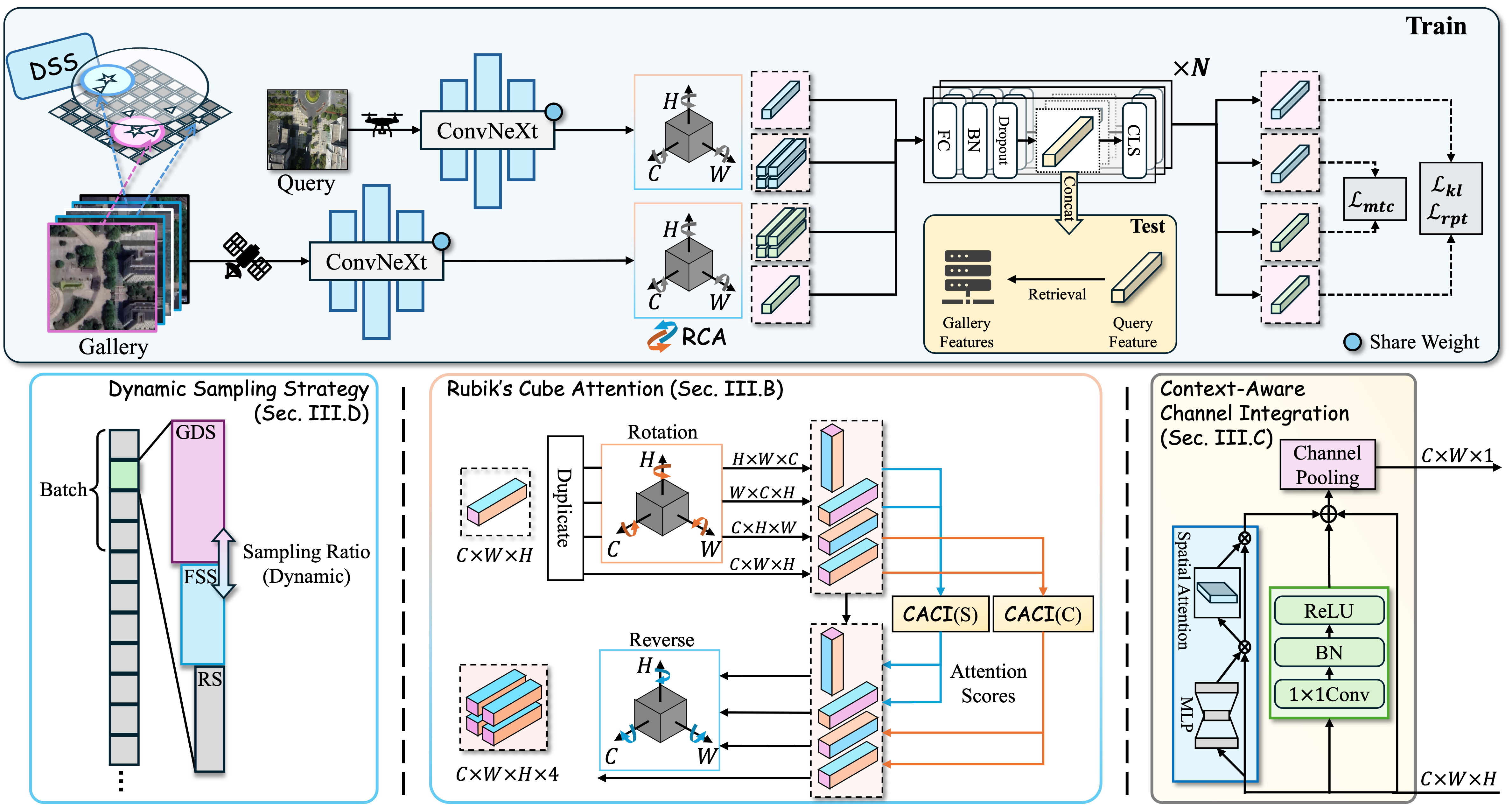}
  \vspace{-15pt}
  \caption{Overview of our CEUSP framework. The framework integrates a Dynamic Sampling Strategy (DSS) to prioritize difficult negative samples during training. The Rubik's Cube Attention (RCA) module, combined with the Context-Aware Channel Integration (CACI) module, captures spatial-channel interactions, with weight sharing applied selectively to enhance performance. During testing, features extracted before the classification layer are matched using cosine similarity.}
  \label{overview}
  \vspace{-10pt}
\end{figure*}

This section firstly provides a detailed overview of the proposed CEUSP framework in Section~\ref{Overview}, which comprising three key components.
The Rubik's Cube Attention (RCA) module (Section~\ref{Rubik's Cube Attention}) enhances feature interaction and fusion, while the Context-Aware Channel Integration (CACI) module (Section~\ref{Context-Aware Channel Integration}) aggregates contextual information across channels, optimizing performance in complex cross-view geo-localization tasks.
The Dynamic Sampling Strategy (DSS) (see Section~\ref{Dynamic Sampling Method}) adjusts sample selection during training to focus on challenging cases, improving the model's robustness.

\subsection{Overview}\label{Overview}  
Cross-view geo-localization aims to map images from different platforms at the same location into similar feature spaces while maintaining distinct representations for images from different locations.
Fig.~\ref{overview} illustrates the CEUSP framework, which integrates a dynamic sampling strategy to prioritize challenging negative samples during training, enhancing model robustness.
ConvNeXt~\cite{liu2022convnet} serves as the primary feature extractor, employing weight-sharing across network branches to leverage geometric and textural similarities between UAV and satellite images.
Following feature extraction, the RCA module, in combination with the CACI module, enables intricate spatial-channel interaction modeling.
Selective weight sharing within certain CACI layers further improves accuracy.
The framework combines representation, metric, and mutual learning to optimize feature representations at category, feature, and distribution levels.
Category-level losses $\mathcal{L}_{rpt}$ are optimized using cross-entropy, while feature-level losses $\mathcal{L}_{mtc}$ employ a Hard-Mining Triplet Loss, calculated after applying global average pooling to the RCA module's output. For distribution-level learning, we minimize the KL divergence $\mathcal{L}_{kl}$ between the predicted distributions of the UAV and satellite branches.
During testing, feature representations prior to the classification layer are used for multi-view comparisons, achieving efficient retrieval.
Detailed descriptions of each module follow in subsequent sections.

\subsection{Rubik's Cube Attention}\label{Rubik's Cube Attention}  
Inspired by the rotational mechanics of a Rubik's Cube, we introduce the Rubik's Cube Attention (RCA) module to enhance cross-dimensional information extraction from input features, as shown in Fig.~\ref{Overview}.
Analogous to a Rubik's Cube's mechanics, where multiple faces undergo manipulation along various axes, generating dynamic positional shifts and interconnections, the RCA module achieves rich and discriminative feature representations by replicating these operations in feature space.

Given an input feature map \( f \in \mathbb{R}^{C \times H \times W} \), where \( C \), \( H \), and \( W \) denote the channel, height, and width dimensions, the RCA module partitions \( f \) into four distinct branches, each undergoing unique operations to improve cross-dimensional interactions.
Let $\mathcal{P}_\sigma$ denote the permutation operation on the dimensions of $f$ according to a specified order $\phi$, $\mathcal{C}$ represent a CACI operation, and $\phi$ the permutation specified for each branch.
For branch \( k \), where \( k \in \{ \text{CHW, HWC, WCH, CWH} \} \), the operations are defined as:

\begin{equation}
  f_k^{\prime} = \mathcal{C}_k(\mathcal{P}_{\phi_k}(f)),
\end{equation}
where \( \mathcal{C}_k \) denotes the CACI applied after permutation \( \phi_k \), rearranging the dimensions in each branch.
To apply attention scaling, a sigmoid activation produces a scaling tensor \( S_k \):
\begin{equation}
  S_k = \sigma(\mathcal{C}_k(\mathcal{P}_{\phi_k}(x))).
\end{equation}
The scaled output for each branch \( k \) is then calculated by element-wise multiplication with the permuted tensor, followed by reverse permutation \( \mathcal{P}_{\phi_k}^{-1} \) to restore the original dimension order:

\begin{equation}
  f_k^{\prime\prime} = \mathcal{P}_{\phi_k}^{-1}(S_k \odot \mathcal{P}_{\phi_k}(f)).
\end{equation}

Finally, the outputs from all branches are aggregated to yield the final output $f^{\prime} = \sum_{k} f_k^{\prime\prime}$.

The rotations in the RCA module promotes both intra- and inter-channel feature propagation, effectively overcoming the spatial constraints of conventional convolution operations.
By leveraging complementary information from spatial and channel dimensions, RCA generates a richer, more comprehensive feature representation, enhancing the model's feature representation capacity.

\subsection{Context-Aware Channel Integration}\label{Context-Aware Channel Integration}  
UAV self-positioning in dense environments, such as in the DenseUAV dataset, requires robust feature extraction to mitigate the effects of environmental variability, including shadows, weather changes, and inconsistent landmarks.
To address these challenges, we introduce the Context-Aware Channel Integration (CACI) module within the RCA framework, designed to enhance discriminative feature extraction across spatial and channel dimensions.
Acting as an attention mechanism, CACI selectively amplifies essential feature channels by generating channel-wise attention weights applied to the rotated feature map, as shown in Figure~\ref{Overview}.

The CACI module combines channel and spatial attention mechanisms, utilizing \(1 \times 1\) convolutions to enable cross-channel interactions.
However, with the feature map rotations introduced by the RCA module, these interactions extend beyond conventional directions, allowing CACI to integrate information more effectively across various spatial and channel dimensions.
This multi-directional integration enhances feature extraction at larger scales.
The channel attention map is computed as:

\begin{equation}
  M_c(X) = \sigma(\text{MLP}(\text{AvgPool}(X))),
\end{equation}
where \(\sigma\) denotes the sigmoid activation function.
The spatial attention mechanism follows the structure of the Convolutional Block Attention Module (CBAM)~\cite{woo2018cbam}, combining average and max-pooling operations to generate the spatial attention map.

Working in tandem with RCA, CACI emphasizes the interaction across spatial dimensions \((W, H)\) and channel dimension \((C)\).
To capture complex relationships across these dimensions efficiently, a weight-sharing strategy is applied for spatial-axis operations and interactions with the input tensor, ensuring consistent processing across all dimensions and enhancing both cohesion and generalization compared to non-weight-sharing methods.

\subsection{Dynamic Sampling Strategy}\label{Dynamic Sampling Method}  
In cross-view geo-localization tasks, including UAV self-positioning, sample selection critically influences model performance.
Building upon dynamic similarity sampling methodologies~\cite{deuser2023sample4geo}, we propose an advanced dataset-based Dynamic Sampling Strategy (DSS).
As illustrated in Figure~\ref{Overview}, the strategy incorporates Geographical Distance Sampling (GDS), Feature Similarities Sampling (FSS), and Random Sampling (RS).
DSS adaptively refines sample selection by balancing geographic relevance and feature diversity within each training batch, thereby enhancing feature matching accuracy and improving model robustness.

Initially, models struggle to differentiate challenging samples due to limited feature understanding.
To address this, we implement a hybrid sampling approach that allocates each batch into three categories:
(1) 50\% of samples are selected based on geographic proximity, directing the model's attention to spatial information by including negative samples with similar GPS coordinates;
(2) 25\% are chosen by cosine similarity to emphasize feature diversity;
and (3) 25\% are randomly sampled to maintain variety and avoid overfitting.
This initial distribution ensures that the model is exposed to a broad spectrum of image perspectives, gradually learning a diverse range of features.

As training progresses, sampling ratios dynamically adjust to reflect the model's evolving feature understanding.
In the later stages, 50\% of samples are selected based on cosine similarity to focus on challenging features, 25\% by geographic proximity to maintain spatial relevance, and 25\% remain randomly sampled for diversity.
This adaptive strategy improves the model's ability to balance geographic context and feature diversity, enhancing robustness and performance in UAV self-positioning tasks.

\subsection{Classifier Head}\label{Prediction Head}  
In cross-view geo-localization tasks, substantial distributional discrepancies between image features from different data acquisition platforms pose significant challenges.
To address these challenges, we employ a set of $N$ classifiers that maps heterogeneous features into a shared feature space, facilitating more reliable comparison and classification.
Each feature vector, after passing through on of these classifiers, is transformed into a class vector, which is used to compute cross-entropy loss during training.
As illustrated in Fig.~\ref{overview}, during the testing phase, the model's feature extraction module produces five sets of feature representations with a dimensionality of $N \times n_b$, where $n_b$ denotes the bottleneck length. $N$ and $n_b$ are set to $5$ and $512$ in this paper.

\subsection{Loss Function}\label{Loss Function}
Our model employs three supervised learning methodologies: representation learning, metric learning, and mutual learning, each calculating losses at the class, feature, and distribution levels.

\textbf{Representation Learning.}
Representation learning optimizes feature extraction for cross-view tasks.
We use cross-entropy loss to train the model to distinguish between classes in feature space, driving the model to learn discriminative representations essential for accurate cross-view image matching.

\textbf{Metric Learning.}
To address distributional discrepancies between features from different platforms in cross-view geographic localization, we use Hard-Mining Triplet Loss. This function maps features to a shared space and minimizes the distance between features with the same geographic label while maximizing the distance between features with different labels. The Hard-Mining Triplet Loss is defined as:
\begin{equation}
  \text{HMTri}(a, p, n) = \max_{(a, p, n)}[\text{Tri}(a, p, n)]
\end{equation}
where
\begin{equation}
  \text{Tri}(a, p, n) = \max (0, D(a, p) - D(a, n) + m)
\end{equation}
Here, \(a\), \(p\), and \(n\) represent anchor, positive, and negative sample features respectively, and \(m\) is the margin. \(D(a, b)\) denotes the cosine similarity between \(a\) and \(b\).
In this study, we set $m=0.3$.

\textbf{Mutual Learning.}
For UAV self-positioning, consistency in class vector distributions across views is critical. Mutual learning, via bidirectional knowledge distillation, minimizes the Kullback-Leibler (KL) divergence between class vector outputs of different views, ensuring similar distributions for same-class images in the shared feature space. The mutual learning loss is:
\begin{equation}
  \text{KLLoss} = \text{KLDiv}(O_d \| O_s) + \text{KLDiv}(O_s \| O_d)
\end{equation}
where \(\text{KLDiv}(O_p \| O_q) = \sum_{i=1}^{N} O_p(i) \times \log\left(\frac{O_p(i)}{O_q(i)}\right)\) calculates the KL divergence between class vectors \(O_p\) and \(O_q\). \(O_d\) and \(O_s\) are the class vector distributions from drone and satellite images, respectively.

To jointly optimize feature discriminability, cross-view alignment, and class distribution consistency, we combine our losses additively.
The total loss is a simple sum of the representation learning, metric learning, and mutual learning losses $\mathcal{L}_{rpt}, \mathcal{L}_{mtc}, \mathcal{L}_{kl}$, enabling simultaneous optimization of these complementary objectives for enhanced cross-view geo-localization performance.

\section{Experimental Results}\label{Experimental Results}
\subsection{Datasets}
\textbf{DenseUAV.}
The DenseUAV dataset consists of densely sampled drone and satellite imagery, emphasizing significant spatial overlap between adjacent frames.
The drone imagery is captured from three distinct altitudes while consistently maintaining identical geographic coordinates under varying temporal and weather conditions.
In addition, the dataset includes 20 levels of satellite images sourced from Google Maps, recorded at different times to facilitate the model's learning of spatial and temporal variations across multiple scales.
This dense sampling introduces challenges for models, particularly in capturing fine-grained details and spatial relationships, enhancing the dataset's relevance for real-world geolocation tasks.
Detailed information about this dataset is provided in Table~\ref{Datasets}.

\textbf{University-1652.} University-1652 is an extensive, cross-view, multi-source dataset designed to address the complexities of cross-platform image matching and geolocation tasks involving UAVs, satellites, and ground cameras.
The dataset's complexity arises from the fact that a single satellite image may correspond to multiple UAV perspectives, presenting a significant challenge for cross-view matching models.
For further details, refer to Table~\ref{Datasets}.

\begin{table}[!t]
  \centering
  \caption{The Data Composition of DenseUAV and University-1652}
  \label{Datasets}
  \resizebox{0.95\linewidth}{!}{
    \begin{tabular}{cccccc}
      \toprule
      \multirow{2}[4]{*}{Subset} & \multicolumn{3}{c}{\#Images} & \multirow{2}[4]{*}{\#Classes} & \multirow{2}[4]{*}{\#Universities}                                 \\
      \cmidrule{2-4}             & UAV                          & Street                        & Satellite                          &      &                        \\
      \midrule
      \multicolumn{6}{c}{DenseUAV}                                                                                                                                   \\
      \midrule
      Training                   & 6768                         & -                             & 13536                              & 2256 & 10                     \\
      Query                      & 2331                         & -                             & 4662                               & 777  & 4                      \\
      Gallery                    & 9099                         & -                             & 18198                              & 3033 & 14                     \\
      \midrule
      \multicolumn{6}{c}{University-1652}                                                                                                                            \\
      \midrule
      Training                   & 37854                        & 11640                         & 701                                & 701  & 33                     \\
      Query                      & 37854                        & 2579                          & 701                                & 701  & \multirow{2}[1]{*}{39} \\
      Gallery                    & 51335                        & 2921                          & 951                                & 951  &                        \\
      \bottomrule
    \end{tabular}%
  }
  \vspace{-5pt}
\end{table}

\subsection{Evaluation Protocols}
\begin{figure}[!t]
  \centering
  \includegraphics[width=1.0\linewidth]{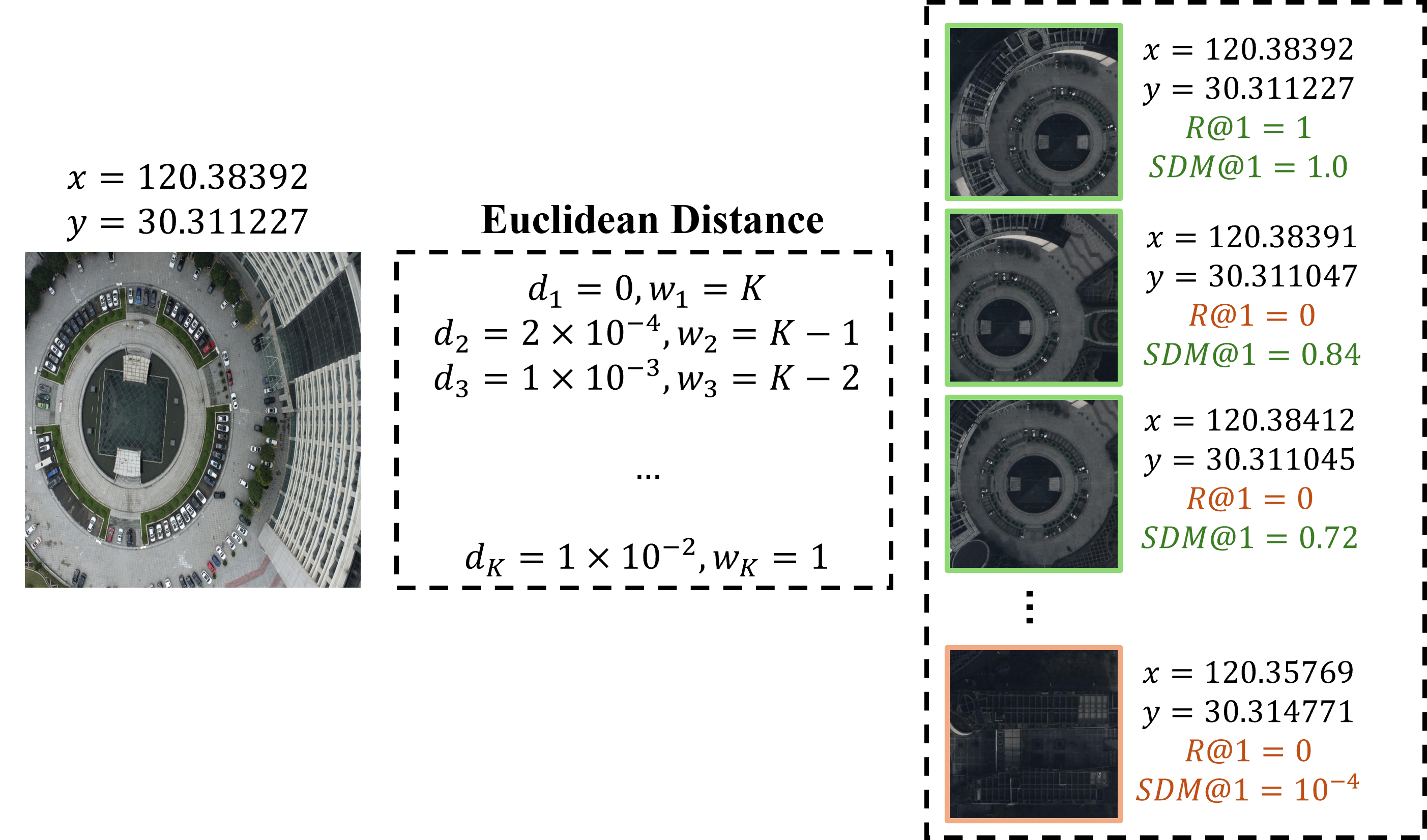}
  \caption{Comparison of R@1 and SDM@K metrics. SDM@K, normalized between 0 and 1, evaluates the spatial Euclidean distance between query and gallery images, assigning higher weights to closer matches. This metric balances retrieval accuracy with spatial precision, tolerating minor deviations while penalizing larger errors, making it particularly suitable for UAV self-positioning tasks.}
  \label{SDM}
  \vspace{-15pt}
\end{figure}

We assess the performance of our model using two established metrics in cross-view geo-localization: Recall@K (R@K) and Average Precision (AP).
In addition to these standard metrics, we introduce a novel evaluation criterion tailored for UAV self-positioning tasks: the Spatial Distance Metric (SDM@K)~\cite{10376356}.
Unlike R@K, which emphasizes exact matches, SDM@K accounts for the spatial continuity inherent in satellite imagery, capturing small positional deviations that frequently occur in UAV operations. These deviations, often due to the dense sampling of satellite data, are inadequately reflected by traditional metrics like R@1, which only assess exact matches. SDM@K mitigates this limitation by incorporating a recall-based measure that considers spatial proximity, allowing minor positional errors to be less severely penalized.
SDM@K is defined as:
\begin{equation}
  \text{SDM@K} = \sum_{i=1}^K \frac{K-i+1}{e^{s\times d_i}} / \sum_{i=1}^K (K-i+1)
\end{equation}
where \( d_i = \sqrt{(x_q - x_i)^2 + (y_q - y_i)^2} \) represents the spatial Euclidean distance between the query and gallery images, and \( (K - i + 1) \) denotes the weight assigned to the \( i \)-th ranked sample. As shown in Fig.~\ref{SDM}, weights are determined by feature distance, with geographically closer gallery images receiving higher weights. \( x_q, y_q \) and \( x_i, y_i \) denote the longitude and latitude of the query and gallery images, respectively, while \( s \) is a scaling factor set to \( 5 \times 10^3 \).

\subsection{Implementation Details}  
We use ConvNeXt-T, pre-trained on ImageNet, as the backbone.
During training, input images are resized to $256 \times 256$ with augmentations, including random padding, cropping, and flipping.
Stochastic gradient descent with a momentum of 0.9 and a weight decay of 0.0005 is applied for optimization. The batch size is set to 32. The learning rate is initialized at 0.003 for backbone parameters and 0.01 for other trainable parameters.
It is reduced by a factor of 0.1 at the 70th and 110th epochs, with training running for 120 epochs in total.
During testing, similarity between query and gallery images is measured using Euclidean distance.
The model is implemented in PyTorch and all experiments are performed on an Nvidia RTX 3090 GPU.

\subsection{Comparison with the State-of-the-art Methods}
\textbf{Results on DenseUAV.}  
The proposed CEUSP framework shows significant improvements in UAV self-positioning, as detailed in Table~\ref{DenseUAV_Results}.
CEUSP achieves a R@1 of 89.45\% and an AP of 79.62\%, outperforming state-of-the-art methods in complex scenarios.
Compared to MCCG, CEUSP improves R@1 by 6.31\% and AP by 7.02\%.
Furthermore, CEUSP surpasses the DenseUAV baseline by 6.44\% in R@1 and 7.52\% in AP, demonstrating superior feature extraction and representation capabilities.

\begin{table*}[!t]
  \centering
  \caption{Comparison with the State-of-the-art Methods on DenseUAV~\cite{10376356} Dataset.
    Top and Second-best Performances are Highlighted in \textbf{Bold} and \underline{Underlined}.
    Input Image Size is $256 \times 256$.}
  \label{DenseUAV_Results}%
  \resizebox{0.95\linewidth}{!}{
    \begin{tabular}{cccccccccc}
      \toprule
      \multicolumn{1}{c}{Methods}              & \multicolumn{1}{c}{Venue} & \multicolumn{1}{c}{Backbone}    & \multicolumn{1}{c}{R@1}             & \multicolumn{1}{c}{R@5}             & \multicolumn{1}{c}{R@top1}           & \multicolumn{1}{c}{AP}              & \multicolumn{1}{c}{SDM@1}           & \multicolumn{1}{c}{SDM@3}           & \multicolumn{1}{c}{SDM@5}           \\
      \midrule
      Triplet Loss~\cite{schroff2015facenet}   & CVPR'15                   & ResNet-50                       & 11.88                               & 32.22                               & -                                    & -                                   & 21.91                               & -                                   & -                                   \\
      Instance Loss~\cite{zheng2020university} & ACMMM'20                  & ResNet-50                       & 13.00                               & 35.78                               & -                                    & -                                   & 23.61                               & -                                   & -                                   \\
      LCM~\cite{ding2020practical}             & Remote Sens'20            & ResNet-50                       & 25.37                               & 50.92                               & -                                    & -                                   & 35.52                               & -                                   & -                                   \\
      MSBA~\cite{zhuang2021faster}             & Remote Sens'21            & ResNet-50                       & 46.13                               & 64.22                               & -                                    & -                                   & 52.64                               & -                                   & -                                   \\
      LPN~\cite{wang2021each}                  & TCSVT'21                  & ResNet-50                       & 32.43                               & 56.80                               & -                                    & -                                   & 40.26                               & -                                   & -                                   \\
      LPN~\cite{wang2021each}                  & TCSVT'21                  & ViT-S                           & 71.77                               & 90.13                               & -                                    & -                                   & 77.95                               & -                                   & -                                   \\
      FSRA~\cite{dai2021transformer}           & TCSVT'21                  & ViT-S                           & 81.21                               & 94.55                               & 99.89                                & 71.93                               & 85.11                               & 83.54                               & 79.74                               \\
      RK-Net~\cite{9779991}                    & TIP'22                    & ResNet-50                       & 38.74                               & 62.85                               & -                                    & -                                   & 45.78                               & -                                   & -                                   \\
      Sample4Geo~\cite{deuser2023sample4geo}                     &      ICCV'23             & ConvNeXt-B                      & {49.38}                   & 78.29                               & 99.40                                & {35.93}                   & 61.72                               & -                               & -                               \\
      {DenseUAV baseline}~\cite{10376356}      & {TIP'23}                  & {ViT-S}                         & {83.01}                             & {{95.58}}                 & {\underline{99.91}}                  & {72.10}                             & {\underline{86.50}}                 & {\underline{84.50}}                 & {\underline{80.44}}                 \\
      MCCG~\cite{10185134}                     & TCSVT'24                  & ConvNeXt-T                      & {83.14}                   & 93.39                               & 99.74                                & \underline{72.60}                   & 85.94                               & 84.32                               & 80.14                               \\
      Yang~\cite{10835173}                     & RA-L'25                  & DINOv2-B                      & \underline{86.27}                   & \textbf{96.83}                               & -                                & {-}                   & \underline{88.87}                               & -                               & -                               \\
      \hdashline
      \cellcolor{gray!10}{CEUSP (Ours)}        & \cellcolor{gray!10}{--}   & \cellcolor{gray!10}{ConvNeXt-T} & \cellcolor{gray!10}{\textbf{89.45}} & \cellcolor{gray!10}{\underline{96.05}} & \cellcolor{gray!10}{\textbf{100.00}} & \cellcolor{gray!10}{\textbf{79.62}} & \cellcolor{gray!10}{\textbf{91.01}} & \cellcolor{gray!10}{\textbf{89.42}} & \cellcolor{gray!10}{\textbf{85.34}} \\
      \bottomrule
    \end{tabular}%
  }
\end{table*}%

\textbf{Results on University-1652.}  
Although primarily designed for UAV self-positioning, CEUSP also performs competitively in traditional cross-view geo-localization tasks.
As shown in Table~\ref{University1652_Results}, in the drone-view target localization task (Drone $\rightarrow$ Satellite) on the University-1652 dataset, CEUSP achieves a R@1 of 90.14\% and an AP of 91.54\%.
In the drone navigation task (Satellite $\rightarrow$ Drone), it attains a R@1 of 93.30\% and an AP of 89.35\%.
Compared to MCCG, CEUSP shows incremental gains of 0.86\% in R@1 and 0.53\% in AP for the UAV-to-Satellite task.
Even against the recent SDPL approach~\cite{chen2024sdpl}, CEUSP remains highly competitive, trailing by only 0.02\% in R@1 and 0.10\% in AP.
These results demonstrate that while CEUSP is optimized for UAV self-positioning task, it performs on par with the latest state-of-the-art methods in traditional cross-view geo-localization tasks.

\begin{table*}[!t]
  \centering
  \caption{Comparison with the State-of-the-art Methods on University-1652~\cite{zheng2020university} Dataset. Top and Second-best Results are Highlighted in \textbf{Bold} and \underline{Underlined}. Although the Dataset Lacks Explicit Geographical Distance Information, the Proposed Model Achieves Competitive Performance.}
  \label{University1652_Results}
  \resizebox{0.95\linewidth}{!}{
    \begin{tabular}{ccccccccccccc}
      \hline
                                               &                       &                         &                       &                                 &                       &                                     &                       & \multicolumn{5}{c}{University-1652}                                                                                                                                                           \\ \cline{9-13}
                                               &                       &                         &                       &                                 &                       &                                     &                       & \multicolumn{2}{c}{Drone-Satellite}    &                                        & \multicolumn{2}{c}{Satellite-Drone}                                                                         \\ \cline{9-10} \cline{12-13}
      \multirow{-3}{*}{Methods}                &                       & \multirow{-3}{*}{Venue} &                       & \multirow{-3}{*}{Backbone}      &                       & \multirow{-3}{*}{Image size}        &                       & Recall@1                               & AP                                     &                                     & Recall@1                     & AP                                     \\
      \hline
      Instance Loss~\cite{zheng2020university} &                       & ACMMM'20                &                       & ResNet-50                       &                       & 512$\times$512                      &                       & 59.69                                  & 64.80                                  &                                     & 73.18                        & 59.40                                  \\
      LCM~\cite{ding2020practical}             &                       & Remote Sens'20          &                       & ResNet-50                       &                       & 512$\times$512                      &                       & 66.65                                  & 70.82                                  &                                     & 79.89                        & 65.38                                  \\
      LPN~\cite{wang2021each}                  &                       & TCSVT'21                &                       & ResNet-50                       &                       & 512$\times$512                      &                       & 77.71                                  & 80.80                                  &                                     & 90.30                        & 78.78                                  \\
      RK-Net~\cite{9779991}                    &                       & TIP'22                  &                       & ResNet-50                       &                       & 512$\times$512                      &                       & 68.10                                  & 71.53                                  &                                     & 80.96                        & 69.35                                  \\
      Sample4Geo~\cite{deuser2023sample4geo}   &                       & ICCV'23                 &                       & ResNet-50                       &                       & 512$\times$512                      &                       & 78.62                                  & 82.11                                  &                                     & 87.45                        & 76.32                                  \\
      AEN(w. LPN)~\cite{10506984}              &                       & SPL'24                  &                       & ResNet-50                       &                       & 256$\times$256                      &                       & 77.40                                  & 80.27                                  &                                     & {90.30}                      & 76.01                                  \\
      \hdashline
      ViT~\cite{dosovitskiy2020image}          &                       & ICLR'20                 &                       & ViT-S                           &                       & 512$\times$512                      &                       & 74.09                                  & 77.82                                  &                                     & 83.31                        & 72.27                                  \\
      FSRA~\cite{dai2021transformer}           &                       & TCSVT'21                &                       & ViT-S                           &                       & 512$\times$512                      &                       & 85.50                                  & 87.53                                  &                                     & {89.73}                      & {84.94}                                \\
      {DenseUAV baseline}~\cite{10376356}      & {}                    & {TIP'23}                & {}                    & {ViT-S}                         & {}                    & {224$\times$224}                    & {}                    & {82.22}                                & {84.78}                                & {}                                  & {87.59}                      & {81.49}                                \\
      TransFG~\cite{10387514}                  &                       & TGRS'24                 &                       & ViT-S                           &                       & 512$\times$512                      &                       & {87.92}                                & {89.99}                                &                                     & {93.37}                      & {87.94}                                \\
      \hdashline
      Swin-B~\cite{liu2021swin}                &                       & CVPR'21                 &                       & Swin-B                          &                       & 256$\times$256                      &                       & 84.15                                  & 86.62                                  &                                     & 90.30                        & 83.55                                  \\
      SwinV2-B~\cite{liu2022swin}              &                       & CVPR'22                 &                       & SwinV2-B                        &                       & 256$\times$256                      &                       & 86.99                                  & 89.02                                  &                                     & 91.16                        & 85.77                                  \\
      F3-Net~\cite{10129939}                   &                       & TGRS'23                 &                       & --                              &                       & 384$\times$384                      &                       & 78.64                                  & 81.60                                  &                                     & --                           & --                                     \\
      Song~\cite{10288351}                     &                       & GRSL'23                 &                       & OSNet                           &                       & 512$\times$512                      &                       & 83.26                                  & 85.84                                  &                                     & 90.30                        & 82.71                                  \\
      Sample4Geo~\cite{deuser2023sample4geo}   &                       & ICCV'23                 &                       & ConvNeXt-B                      &                       & 512$\times$512                      &                       & \textbf{92.65}                                & \textbf{93.81}                                &                                     & \textbf{95.14}                      & \textbf{91.39}                                \\
      MFJRLN(Lcro)~\cite{10517694}             &                       & TGRS'24                 &                       & Swin-B                          &                       & 224$\times$224                      &                       & 87.61                                  & 89.57                                  &                                     & 91.07                        & 86.72                                  \\
      GeoFormer~\cite{10506965}                &                       & J-STARS'24              &                       & E-Swin-B                        &                       & 224$\times$224                      &                       & {88.16}                                & {90.03}                                &                                     & {91.87}                      & {87.92}                                \\
      MCCG~\cite{10185134}                     &                       & TCSVT'24                &                       & ConvNeXt-T                      &                       & 256$\times$256                      &                       & {89.28}                                & {91.01}                                &                                     & \underline{94.29}               & {89.29}                                \\
      SDPL~\cite{chen2024sdpl}                 &                       & TCSVT'24                &                       & SwinV2-B                        &                       & 256$\times$256                      &                       & \underline{90.16}                         & \underline{91.64}                         &                                     & {93.58}            & \underline{89.45}                         \\

      \hdashline
      \cellcolor{gray!10}{CEUSP (Ours)}        & \cellcolor{gray!10}{} & \cellcolor{gray!10}{--} & \cellcolor{gray!10}{} & \cellcolor{gray!10}{ConvNeXt-T} & \cellcolor{gray!10}{} & \cellcolor{gray!10}{256$\times$256} & \cellcolor{gray!10}{} & \cellcolor{gray!10}{${90.14}$} & \cellcolor{gray!10}{${91.54}$} & \cellcolor{gray!10}{}               & \cellcolor{gray!10}{${93.30}$} & \cellcolor{gray!10}{${89.35}$} \\
      \hline
    \end{tabular}
  }
\end{table*}

\subsection{Ablation Studies}
\begin{figure}[!t]
  \centering
  \includegraphics[width=0.9\linewidth]{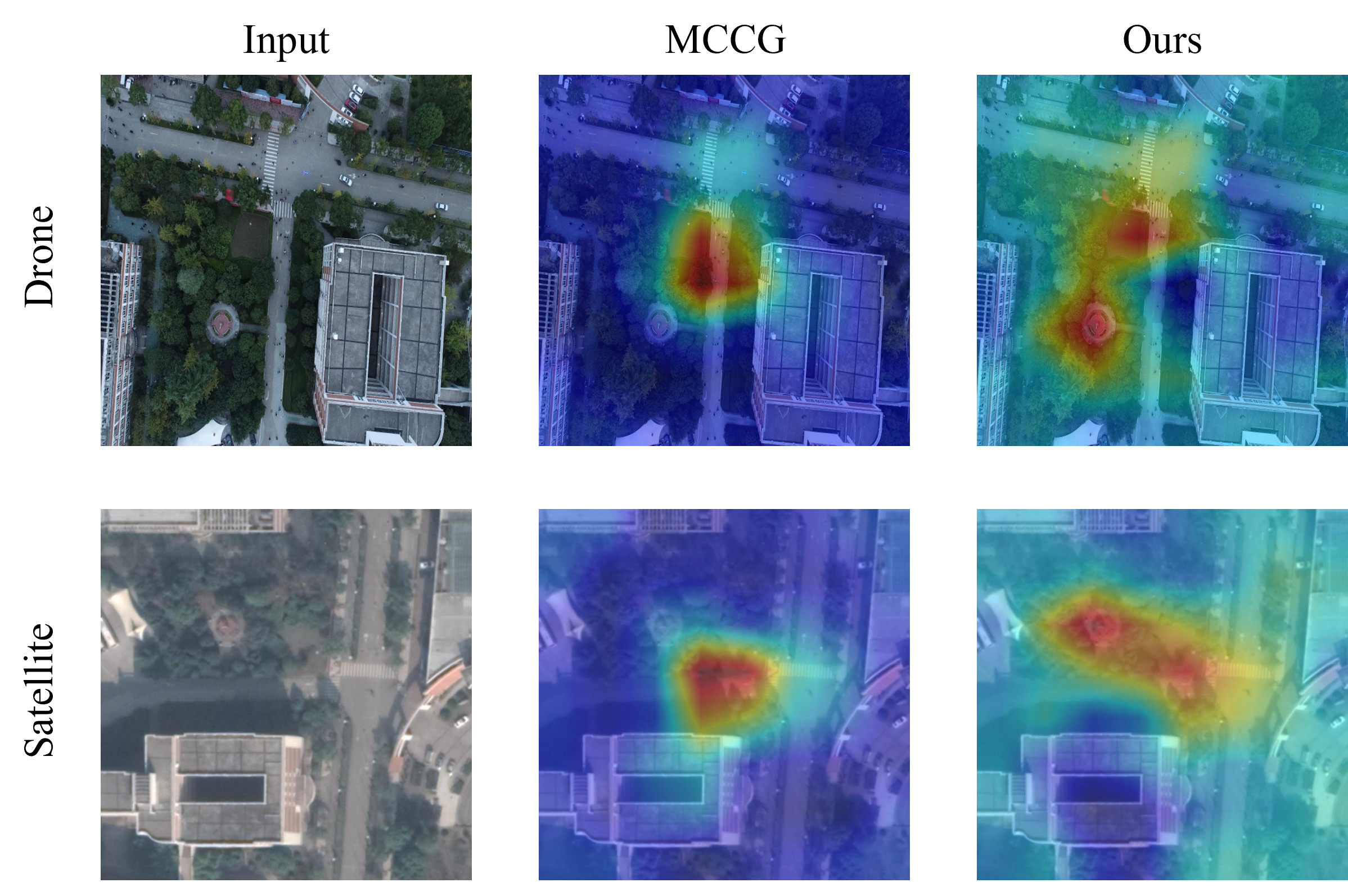}
  \caption{Heatmaps generated by MCCG and CEUSP. CEUSP shows a greater focus on significant landmarks, while MCCG emphasizes central features. CEUSP effectively highlights critical elements, such as the pavilion and central intersection, demonstrating enhanced scene comprehension.}
  \label{heatmaps_compare}
  \vspace{-15pt}
\end{figure}
The primary experiments are conducted on the DenseUAV dataset.

\textbf{Impact of Backbone Network Architecture.}  
We evaluate the performance of various backbone networks, including ViT-S, ResNet-50, ConvNeXt-T, and ConvNeXt-S, on UAV self-positioning tasks (Table~\ref{backbone_network}).
ConvNeXt-T shows better performance, achieving a R@1 of 89.45\% and an AP of 79.62\%, and demonstrates stronger feature extraction in SDM@1 and SDM@5.
Although ViT-S performs reasonably well, it falls short compared to the ConvNeXt variants despite its global attention mechanism.
Experiment shows ConvNeXt-S, with its deeper architecture, underperforms slightly relative to ConvNeXt-T, possibly due to redundant feature extraction.

\begin{table}[!t]
  \centering
  \caption{Comparison of RestNet, ViT and ConvNeXt Networks for UAV Self-Positioning Tasks on DenseUAV}
  \label{backbone_network}%
  \resizebox{1.0\linewidth}{!}{
    \begin{tabular}{cccccc}
      \toprule
      \multicolumn{1}{c}{Backbone} & \multicolumn{1}{c}{R@1} & \multicolumn{1}{c}{R@5} & \multicolumn{1}{c}{AP} & \multicolumn{1}{c}{SDM@1} & \multicolumn{1}{c}{SDM@5} \\
      \midrule
      ResNet-50                    & 38.78                   & 63.15                   & 26.49                  & 47.59                     & 40.43                     \\
      ViT-S                        & 84.94                   & 96.31                   & 76.31                  & 87.50                     & 83.07                     \\
      ConvNeXt-T                   & \textbf{89.45}          & \textbf{96.05}          & \textbf{79.62}         & \textbf{91.01}            & \textbf{85.34}            \\
      ConvNeXt-S                   & 87.17                   & 95.07                   & 78.19                  & 89.14                     & 84.10                     \\
      \bottomrule
    \end{tabular}%
  }
  \vspace{-15pt}
\end{table}%

\textbf{Influence of Dynamic Sampling Strategy.}  
We conduct a series of experiments to assess the effectiveness of the DSS (Table~\ref{dynamic_sampling_ratio}).
Random sampling (RS) achieves a R@1 of 85.84\%, while geographical distance sampling (GDS) shows a lower performance at 52.72\%.
In contrast, feature similarity sampling (FSS) showed stronger performance.
The integration of GDS and FSS in mixed sampling yields R@1 values between 88.72\% and 89.45\%, indicating that the performance improvements result from combining strategies rather than specific sampling ratios.
The relative insensitivity of performance to these ratios demonstrates the robustness of the proposed method.

\begin{table}[!t]
  \centering
  \caption{Comparative Analysis of Sampling Strategies and Mixed Sampling Ratios in UAV Self-positioning Tasks. RS Denotes Random Sampling, GDS Represents Geographical Distance Sampling, and FSS Indicates Feature Similarities Sampling}
  \label{dynamic_sampling_ratio}%
  \begin{tabular}{cccccc}
    \toprule
    \multicolumn{1}{c}{Sampling Strategy} & \multicolumn{1}{c}{R@1} & \multicolumn{1}{c}{R@5} & \multicolumn{1}{c}{AP} & \multicolumn{1}{c}{SDM@1} & \multicolumn{1}{c}{SDM@5} \\
    \midrule
    RS Only                               & 85.84                   & 96.10                   & 77.00                  & 88.24                     & 83.82                     \\
    GDS Only                              & 52.72                   & 73.14                   & 40.93                  & 56.05                     & 49.89                     \\
    FSS Only                              & 88.37                   & 95.97                   & 80.16                  & 90.10                     & 85.59                     \\
    GDS:FSS = 1:2                         & \textbf{89.45}          & 96.05                   & 79.62                  & 91.01                     & 85.34                     \\
    GDS:FSS = 2:1                         & 88.72                   & 95.37                   & 80.12                  & 90.56                     & 85.84                     \\
    GDS:FSS = 1:1                         & 89.15                   & \textbf{96.78}          & \textbf{80.41}         & \textbf{91.09}            & \textbf{86.37}            \\
    \bottomrule
  \end{tabular}%
  \vspace{-15pt}
\end{table}%

Further experiments explore the impact of starting mixed sampling at different epochs (Table~\ref{dynamic_sampling_sim_sample_start_epoch}) and the frequency of feature similarity calculations (Table~\ref{dynamic_sampling_cal_sim_every_n_epoch}).
Delaying mixed sampling until epoch 20 achieves the highest R@1, though the variations across epochs are minimal.
Similarly, adjusting the frequency of similarity calculations causes only minor performance fluctuations, with overly frequent calculations disrupting learning dynamics.
These results suggest that delaying the initiation of mixed sampling may offer benefits by allowing the model to first stabilize its understanding of the data.
However, the minimal performance differences across starting epochs indicate that the timing of mixed sampling has limited influence on overall effectiveness.
This reinforces the conclusion that the robustness of mixed sampling arises from its adaptability rather than the precision of its parameter implementation.

\begin{table}[!t]
  \centering
  \caption{Performance Metrics for Varying Initiation Epochs of Mixed Sampling Strategy}
  \label{dynamic_sampling_sim_sample_start_epoch}%
  \begin{tabular}{cccccc}
    \toprule
    \multicolumn{1}{c}{Sampling Strategy} & \multicolumn{1}{c}{R@1} & \multicolumn{1}{c}{R@5} & \multicolumn{1}{c}{AP} & \multicolumn{1}{c}{SDM@1} & \multicolumn{1}{c}{SDM@5} \\
    \midrule
    0                                     & 88.46                   & 95.69                   & 79.94                  & 90.22                     & 86.01                     \\
    5                                     & 88.37                   & \textbf{96.61}          & \textbf{80.03}         & 90.54                     & 86.14                     \\
    10                                    & 88.85                   & 95.74                   & 79.32                  & 90.76                     & \textbf{86.17}            \\
    20                                    & \textbf{89.45}          & 96.05                   & 79.62                  & \textbf{91.01}            & 85.34                     \\
    \bottomrule
  \end{tabular}%
\end{table}%

\begin{table}[!t]
  \centering
  \caption{Performance Metrics for Different Intervals of Feature Similarity Calculation in Mixed Sampling Strategy}
  \label{dynamic_sampling_cal_sim_every_n_epoch}%
  \begin{tabular}{cccccc}
    \toprule
    \multicolumn{1}{c}{Sampling Strategy} & \multicolumn{1}{c}{R@1} & \multicolumn{1}{c}{R@5} & \multicolumn{1}{c}{AP} & \multicolumn{1}{c}{SDM@1} & \multicolumn{1}{c}{SDM@5} \\
    \midrule
    2                                     & 88.37                   & \textbf{97.08}          & 79.51                  & 90.42                     & 85.77                     \\
    4                                     & 88.85                   & 96.74                   & \textbf{80.32}         & 90.77                     & \textbf{86.15}            \\
    6                                     & \textbf{89.45}          & 96.05                   & 79.62                  & \textbf{91.01}            & 85.34                     \\
    8                                     & 87.22                   & 95.11                   & 78.19                  & 89.39                     & 84.40                     \\
    \bottomrule
  \end{tabular}%
\end{table}%

\textbf{Effectiveness of RCA and CACI Modules.}  
Table~\ref{module} shows the contributions of the RCA and CACI modules to overall model performance.
Removing the RCA module causes a noticeable reduction in R@1 and AP, while omitting the CACI module results in a similar performance drop.
Excluding both modules leads to a substantial decline, with R@1 and AP falling to 74.77\% and 62.41\%, respectively.
These findings highlight the crucial role of RCA and CACI in feature extraction and representation.
In contrast, the full CEUSP model achieves better performance, demonstrating the synergistic effect of these modules.

\begin{table}[!t]
  \centering
  \caption{Impact of RCA and CACI Modules on Model Performance Metrics in UAV Self-Positioning Tasks}
  \label{module}
  \resizebox{1.0\linewidth}{!}{
    \begin{tabular}{cccccc}
      \toprule
      \multicolumn{1}{c}{Method} & \multicolumn{1}{c}{R@1} & \multicolumn{1}{c}{R@5} & \multicolumn{1}{c}{AP} & \multicolumn{1}{c}{SDM@1} & \multicolumn{1}{c}{SDM@5} \\
      \midrule
      w/o CACI                   & 88.42                   & 95.92                   & 78.59                  & 90.37                     & {86.00}                   \\
      w/o RCA                    & 87.50                   & 95.10                   & 78.13                  & 89.32                     & 84.53                     \\
      w/o RCA and CACI           & 74.77                   & 89.45                   & 62.41                  & 77.21                     & 70.62                     \\
      CEUSP (Ours)               & \textbf{89.45}          & \textbf{96.05}          & \textbf{79.62}         & \textbf{91.01}            & \textbf{85.34}            \\
      \bottomrule
    \end{tabular}
  }
\end{table}

\textbf{Sensitivity to Input Image Size.}  
Table~\ref{img_size} summarizes the effect of varying input image sizes on model performance.
Increasing the input size from $224$ to $256$ pixels yields relatively better results, with R@1 and AP scores highest at $256$.
Although larger image sizes theoretically offer more spatial information, the model's capacity to effectively leverage this additional data appears constrained, possibly due to overfitting or the complexity of processing higher-resolution inputs.
Thus, $256$ pixels proves to be an effective input resolution across all key evaluation metrics.

\begin{table}[!t]
  \centering
  \caption{\cq{Performance comparison of CEUSP with different input sizes on DenseUAV.}}
  \label{img_size}
  \resizebox{0.9\linewidth}{!}{
    \begin{tabular}{cccccc}
      \toprule
      \multicolumn{1}{c}{Image Size} & \multicolumn{1}{c}{R@1} & \multicolumn{1}{c}{R@5} & \multicolumn{1}{c}{AP} & \multicolumn{1}{c}{SDM@1} & \multicolumn{1}{c}{SDM@5} \\
      \midrule
      $224 \times 224$               & 88.07                   & 95.71                   & 78.86                  & 89.87                     & 84.77                     \\
      $256 \times 256$               & \textbf{88.72}          & 95.84                   & \textbf{79.97}         & \textbf{90.46}            & 85.47                     \\
      $320 \times 320$               & 88.25                   & \textbf{96.65}          & 79.81                  & 90.10                     & \textbf{85.59}            \\
      $384 \times 384$               & 84.13                   & 95.15                   & 76.65                  & 86.84                     & 83.11                     \\
      $512 \times 512$               & 86.27                   & 95.45                   & 76.52                  & 88.42                     & 83.25                     \\
      \bottomrule
    \end{tabular}
  }
\end{table}

\textbf{Weight Sharing in the CACI Module.}  
An experiment was conducted to assess the impact of weight sharing within the CACI module.
Results in Table~\ref{sharing_weights} indicate that the weight-sharing configuration performs better than the non-sharing approach across metrics.
Specifically, R@1, R@5, and AP values improved significantly, suggesting enhanced feature representation and retrieval accuracy.

\begin{table}[!t]
  \centering
  \caption{Performance Comparison of Weight Sharing and Non-Sharing Configurations in the CACI Module}
  \label{sharing_weights}
  \resizebox{0.9\linewidth}{!}{
    \begin{tabular}{cccccc}
      \toprule
      \multicolumn{1}{c}{Method} & \multicolumn{1}{c}{R@1} & \multicolumn{1}{c}{R@5} & \multicolumn{1}{c}{AP} & \multicolumn{1}{c}{SDM@1} & \multicolumn{1}{c}{SDM@5} \\
      \midrule
      Not Sharing                & 86.66                   & 95.20                   & 77.36                  & 88.92                     & 83.63                     \\
      Sharing                    & \textbf{89.45}          & \textbf{96.05}          & \textbf{79.62}         & \textbf{91.01}            & \textbf{85.34}            \\
      \bottomrule
    \end{tabular}
  }
\end{table}

\textbf{Evaluation on Position Shifting.}  
\begin{table*}[!t]
  \centering
  \caption{Performance Evaluation of FSRA, MCCG and CEUSP Under Position Shifting on the DenseUAV Dataset using Black Pad and Flip Pad. Top and Second-best Performances are Highlighted in Red and Blue.}
  \label{position_shifting_DenseUAV}%
  \resizebox{1.0\linewidth}{!}{
    \begin{tabular}{cccccccccccccccccc}
      \toprule
      \multirow{3}[6]{*}{Padding Pixel}                                                          & \multicolumn{8}{c}{Black Pad}        &                                      & \multicolumn{8}{c}{Flip Pad}                                                                                                                                                                                                                                                                                                                                                                                                                                                         \\
      \cmidrule{2-9}\cmidrule{11-18}                                                             & \multicolumn{2}{c}{FSRA}             &                                      & \multicolumn{2}{c}{MCCG}     &                                      & \multicolumn{2}{c}{CEUSP (Ours)}     &  & \multicolumn{2}{c}{FSRA}                          &                                      & \multicolumn{2}{c}{MCCG} &                                       & \multicolumn{2}{c}{CEUSP (Ours)}                                                                                                                                                                       \\
      \cmidrule{2-3}\cmidrule{5-6}\cmidrule{8-9}\cmidrule{11-12}\cmidrule{14-15}\cmidrule{17-18} & AP                                   & SDM@1                                &                              & AP                                   & SDM@1                                &  & AP                                                & SDM@1                                &                          & AP                                    & SDM@1                                &  & AP                                   & SDM@1                                &  & AP                                   & SDM@1                                \\
      \midrule
      0                                                                                          & {${71.93}_{-0}$    }                 & {${85.11}_{-0}$    }                 &                              & \textcolor{blue}{${72.60}_{-0}$}     & \textcolor{blue}{${85.94}_{-0}$}     &  & \textcolor{red}{${79.62}_{-0}$}                   & \textcolor{red}{${91.01}_{-0}$    }  &                          & {${71.93}_{-0}$ }                     & {${85.11}_{-0}$}                     &  & \textcolor{blue}{${72.60}_{-0}$}     & \textcolor{blue}{${85.94}_{-0}$}     &  & \textcolor{red}{${79.62}_{-0}$}      & \textcolor{red}{${91.01}_{-0}$}      \\
      10                                                                                         & {${70.17}_{-1.76}$ }                 & {${83.71}_{-1.40}$ }                 &                              & \textcolor{blue}{${71.05}_{-1.55}$}  & \textcolor{blue}{${85.06}_{-0.88}$}  &  & \textcolor{red}{${78.66}_{-0.96}$              }  & \textcolor{red}{${90.41}_{-0.60}$ }  &                          & {${69.61}_{-2.32}$ }                  & {${83.25}_{-1.86}$}                  &  & \textcolor{blue}{${70.63}_{-1.97}$}  & \textcolor{blue}{${84.64}_{-1.30}$}  &  & \textcolor{red}{${77.54}_{-2.08}$}   & \textcolor{red}{${89.52}_{-1.49}$}   \\
      20                                                                                         & \textcolor{blue}{${59.08}_{-12.85}$} & \textcolor{blue}{${75.17}_{-9.94}$ } &                              & {${58.22}_{-14.38}$}                 & {${74.33}_{-11.61}$}                 &  & \textcolor{red}{${67.49}_{-12.13}$             }  & \textcolor{red}{${82.14}_{-8.87}$ }  &                          & \textcolor{blue}{${58.00}_{-13.93}$ } & \textcolor{blue}{${74.53}_{-10.58}$} &  & {${57.05}_{-15.55}$}                 & {${72.96}_{-12.98}$}                 &  & \textcolor{red}{${62.69}_{-16.93}$}  & \textcolor{red}{${78.26}_{-12.75}$}  \\
      30                                                                                         & \textcolor{blue}{${40.31}_{-31.62}$} & \textcolor{blue}{${58.68}_{-26.43}$} &                              & {${37.81}_{-34.79}$}                 & {${57.57}_{-28.37}$}                 &  & \textcolor{red}{${45.43}_{-34.19}$             }  & \textcolor{red}{${63.37}_{-27.64}$}  &                          & \textcolor{red}{${38.39}_{-33.54}$ }  & \textcolor{red}{${57.40}_{-27.71}$}  &  & {${36.04}_{-36.56}$}                 & {${55.74}_{-30.20}$}                 &  & \textcolor{blue}{${37.65}_{-41.97}$} & \textcolor{blue}{${56.60}_{-34.41}$} \\
      40                                                                                         & \textcolor{red}{${22.50}_{-49.43}$}  & \textcolor{red}{${44.01}_{-41.10}$}  &                              & {${19.64}_{-52.96}$}                 & \textcolor{blue}{${42.83}_{-43.11}$} &  & \textcolor{blue}{${20.19}_{-59.43}$             } & {${42.01}_{-49.00}$}                 &                          & \textcolor{red}{${21.24}_{-50.69}$ }  & \textcolor{red}{${43.45}_{-41.66}$}  &  & \textcolor{blue}{${18.63}_{-53.97}$} & \textcolor{blue}{${41.61}_{-44.33}$} &  & {${18.01}_{-61.61}$}                 & {${40.58}_{-}50.43$}                 \\
      50                                                                                         & \textcolor{red}{${12.12}_{-59.81}$}  & \textcolor{red}{${35.78}_{-49.33}$}  &                              & \textcolor{blue}{${9.43 }_{-63.71}$} & {${34.87}_{-51.07}$}                 &  & {${9.13 }_{-70.49}$             }                 & \textcolor{blue}{${35.01}_{-56.00}$} &                          & \textcolor{red}{${11.53}_{-60.10}$ }  & \textcolor{red}{${36.12}_{-48.99}$}  &  & {${8.93 }_{-63.67}$}                 & {${34.20}_{-51.74}$}                 &  & \textcolor{blue}{${9.02 }_{-70.60}$} & \textcolor{blue}{${35.31}_{-55.70}$} \\
      60                                                                                         & \textcolor{red}{${6.69 }_{-65.24}$}  & \textcolor{red}{${31.71}_{-53.40}$}  &                              & {${5.39 }_{-67.21}$}                 & {${31.07}_{-54.87}$}                 &  & \textcolor{blue}{${5.45 }_{-74.17}$             } & \textcolor{blue}{${32.46}_{-58.55}$} &                          & \textcolor{red}{${6.85 }_{-65.08}$ }  & \textcolor{blue}{${31.89}_{-53.22}$} &  & {${5.24 }_{-67.36}$}                 & {${30.84}_{-55.10}$}                 &  & \textcolor{blue}{${5.46 }_{-74.16}$} & \textcolor{red}{${32.21}_{-58.80}$}  \\
      \bottomrule
    \end{tabular}%
  }
\end{table*}%
\begin{table*}[!t]
  \centering
  \caption{Performance Evaluation of FSRA, MCCG and CEUSP Under Position Shifting on the University-1652 Dataset using Black Pad and Flip Pad. Top and Second-best Performances are Highlighted in Red and Blue.}
  \label{position_shifting_University}%
  \resizebox{1.0\linewidth}{!}{
    \begin{tabular}{cccccccccccccccccc}
      \toprule
      \multirow{3}[6]{*}{Padding Pixel}                                                          & \multicolumn{8}{c}{Black Pad}        &       & \multicolumn{8}{c}{Flip Pad}                                                                                                                                                                                                                                                                                                                                    \\
      \cmidrule{2-9}\cmidrule{11-18}                                                             & \multicolumn{2}{c}{FSRA}             &       & \multicolumn{2}{c}{MCCG}     &                                     & \multicolumn{2}{c}{CEUSP (Ours)} &  & \multicolumn{2}{c}{FSRA}             &       & \multicolumn{2}{c}{MCCG} &                                      & \multicolumn{2}{c}{CEUSP (Ours)}                                                                                                    \\
      \cmidrule{2-3}\cmidrule{5-6}\cmidrule{8-9}\cmidrule{11-12}\cmidrule{14-15}\cmidrule{17-18} & AP                                   & SDM@1 &                              & AP                                  & SDM@1                            &  & AP                                   & SDM@1 &                          & AP                                   & SDM@1                            &  & AP                                  & SDM@1 &  & AP                                   & SDM@1 \\
      \midrule
      0                                                                                          & {${84.77}_{-0}$}                     & -     &                              & \textcolor{blue}{${91.01}_{-0}$}    & -                                &  & \textcolor{red}{${91.54}_{-0}$}      & -     &                          & {${84.77}_{-0}$}                     & -                                &  & \textcolor{blue}{${91.01}_{-0}$}    & -     &  & \textcolor{red}{${91.54}_{-0}$}      & -     \\
      10                                                                                         & {${84.13}_{-0.64}$}                  & -     &                              & \textcolor{blue}{${90.95}_{-0.06}$} & -                                &  & \textcolor{red}{${91.01}_{-0.53}$}   & -     &                          & {${84.19}_{-0.58}$}                  & -                                &  & \textcolor{blue}{${90.14}_{-0.87}$} & -     &  & \textcolor{red}{${90.77}_{-0.77}$}   & -     \\
      20                                                                                         & {${82.70}_{-2.07}$}                  & -     &                              & \textcolor{red}{${90.51}_{-0.50}$}  & -                                &  & \textcolor{blue}{${89.66}_{-1.88}$}  & -     &                          & {${82.26}_{-2.51}$}                  & -                                &  & \textcolor{blue}{${88.42}_{-2.59}$} & -     &  & \textcolor{red}{${88.65}_{-2.89}$}   & -     \\
      30                                                                                         & {${80.03}_{-4.74}$}                  & -     &                              & \textcolor{red}{${89.61}_{-1.40}$}  & -                                &  & \textcolor{blue}{${86.70}_{-4.84}$}  & -     &                          & {${78.46}_{-6.31}$}                  & -                                &  & \textcolor{blue}{${85.52}_{-5.49}$} & -     &  & \textcolor{blue}{${84.05}_{-7.49}$}  & -     \\
      40                                                                                         & {${76.41}_{-8.36}$}                  & -     &                              & \textcolor{red}{${87.70}_{-3.31}$}  & -                                &  & \textcolor{blue}{${81.00}_{-10.54}$} & -     &                          & {${73.13}_{-11.64}$}                 & -                                &  & \textcolor{red}{${81.18}_{-9.83}$}  & -     &  & \textcolor{blue}{${75.95}_{-15.59}$} & -     \\
      50                                                                                         & \textcolor{blue}{${71.60}_{-13.17}$} & -     &                              & \textcolor{red}{${84.69}_{-6.32}$}  & -                                &  & {${71.20}_{-20.34}$}                 & -     &                          & \textcolor{blue}{${66.07}_{-18.70}$} & -                                &  & \textcolor{red}{${75.03}_{-15.98}$} & -     &  & {${63.91}_{-27.63}$}                 & -     \\
      60                                                                                         & \textcolor{blue}{${65.76}_{-19.01}$} & -     &                              & \textcolor{red}{${80.70}_{-10.31}$} & -                                &  & {${59.97}_{-31.57}$}                 & -     &                          & \textcolor{blue}{${57.96}_{-26.81}$} & -                                &  & \textcolor{red}{${67.50}_{-23.51}$} & -     &  & {${51.11}_{-40.43}$}                 & -     \\
      \bottomrule
    \end{tabular}%
  }
\end{table*}%
We evaluate the robustness of CEUSP against position shifting using two padding strategies: Black Pad and Flip Pad.
As shown in Table~\ref{position_shifting_DenseUAV}, CEUSP attains higher AP and SDM@1 scores than FSRA and MCCG when Padding Pixel values are minimal (0 to 20).
However, a notable decline in CEUSP's performance is observed as padding exceeds 40 pixels.
This suggests that CEUSP effectively captures dense features when spatial integrity is preserved, leveraging fine-grained details essential for accurate feature localization.
As padding size increases, spatial relationships among features become more distorted, resulting in reduced feature coherence.
Thus, while CEUSP performs well under minimal perturbation, larger deviations may limit its ability to align and extract meaningful features accurately, affecting its overall effectiveness.

Additionally, for a thorough comparison, we evaluate the University-1652 dataset within the cross-view geo-localization task.
As shown in Table~\ref{position_shifting_University}, the results align closely with those on DenseUAV, further validating the properties and effectiveness of the proposed CEUSP method.

\begin{figure}[!t]
  \centering
  \includegraphics[width=\linewidth]{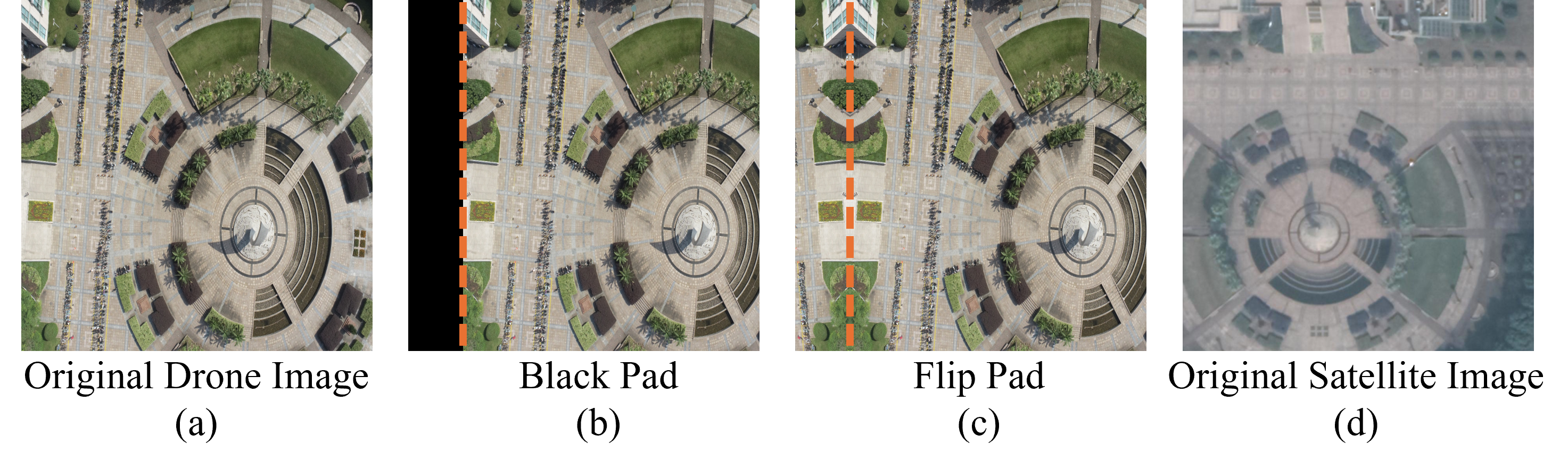}
  \caption{Illustration of Black Pad and Flip Pad methods. (a) Original drone image. (b) Image with a black padding block on the left and a corresponding strip removed from the right. (c) Image created by mirroring a strip on the left and cropping a strip from the right. (d) Original satellite image.}
  \label{position_shifting}
  \vspace{-15pt}
\end{figure}

\subsection{Qualiative Results}  

\begin{figure*}[!t]
  \centering
  \includegraphics[width=0.95\linewidth]{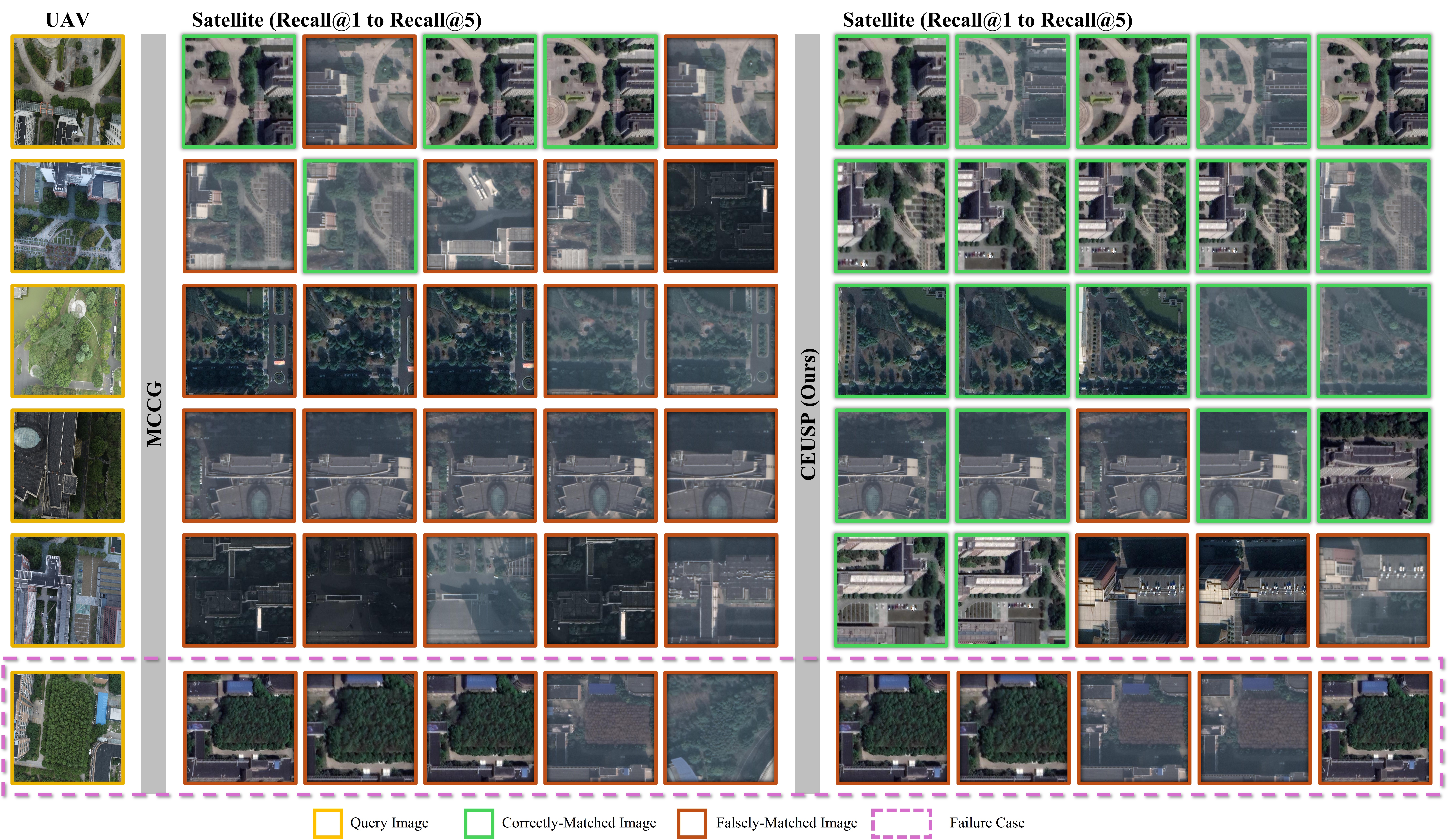}
  \caption{Image retrieval comparison for UAV self-positioning tasks using the MCCG and CEUSP frameworks. Incorrectly matched images are marked with red bounding boxes. Note that dense sampling across varying altitudes in the DenseUAV dataset may result in small deviations that yield incorrect matches despite visual similarity.}
  \label{image_retrieval_result}
  \vspace{-15pt}
\end{figure*}

To evaluate the proposed method qualitatively, we visualize retrieval results from our approach alongside those from the state-of-the-art method, as shown in Fig.\ref{image_retrieval_result}.
For simpler query scenes, both methods retrieve accurate and relevant images (the first set in Fig.~\ref{image_retrieval_result}).
However, as scene complexity increases--with densely clustered buildings or less prominent target objects--our RCA module extracts more distinctive feature representations, yielding relatively better retrieval accuracy (the second and third sets in Fig.~\ref{image_retrieval_result}).

The DenseUAV dataset's dense sampling adds challenges, even for straightforward scenes, where subtle differences in image features can cause retrieval deviations, a difficulty that compounds with increased scene complexity.
Comparatively, our proposed method still maintains robust R@1 accuracy even in complex scenarios  (the fourth and fifth sets of Fig.~\ref{image_retrieval_result}).
Failure cases appear in the sixth set of Fig.~\ref{image_retrieval_result}, where both our method and others struggle to localize UAVs accurately in scenes with major landscape changes cause by time.

Additionally, heatmap visualizations generated by MCCG and our CEUSP model provide further insight into feature focus from both UAV and satellite perspectives, as shown in Fig.~\ref{heatmaps_compare}.
The CEUSP model, using fine-grained features from the RCA module, demonstrates a stronger ability to emphasize multiple relevant buildings and spatial features within scenes.
In contrast, MCCG tends to focus centrally, while CEUSP highlights significant information, such as the pavilion and central intersection, indicating a more comprehensive understanding of the scene.

\subsection{Limitations}  
Ablation studies indicate that increasing model depth does not consistently yield better performance, as ConvNeXt-T performs better than deeper architectures.
This may result from CEUSP's limited capacity for diverse feature capture, with deeper models potentially introducing redundant information that weakens global representations.
Additionally, CEUSP is sensitive to spatial distortions; larger positional shifts disrupt spatial coherence, affecting feature alignment and performance.
Although CEUSP remains robust under minor perturbations, its effectiveness declines with significant displacements.

\section{Conclusion}\label{Conclusion}  
In this paper, we present CEUSP, a novel method tailored for UAV self-positioning tasks.
Built on a ConvNeXt-based architecture, CEUSP integrates a Dynamic Sampling Strategy (DSS) with the Rubik's Cube Attention (RCA) module to autonomously capture and reconfigure significant information, enhancing localization accuracy, especially in dense urban settings.
Experimental evaluations show that CEUSP achieves competitive results on both the DenseUAV dataset for UAV self-positioning and the University-1652 dataset for traditional cross-view geo-localization tasks, highlighting its adaptability for broader geographic localization applications.
Ablation studies confirm that CEUSP effectively mitigates positional biases and scale variations, promoting robustness in complex environments.
Future work will focus on architectural refinements to extend CEUSP's high performance to more challenging, dynamic scene cross-view matching tasks, thereby advancing UAV self-positioning accuracy across diverse, real-world scenarios.


\bibliographystyle{IEEEtran}
\bibliography{reference.bib}

\begin{thebibliography}{10}
\providecommand{\url}[1]{#1}
\csname url@samestyle\endcsname
\providecommand{\newblock}{\relax}
\providecommand{\bibinfo}[2]{#2}
\providecommand{\BIBentrySTDinterwordspacing}{\spaceskip=0pt\relax}
\providecommand{\BIBentryALTinterwordstretchfactor}{4}
\providecommand{\BIBentryALTinterwordspacing}{\spaceskip=\fontdimen2\font plus
\BIBentryALTinterwordstretchfactor\fontdimen3\font minus \fontdimen4\font\relax}
\providecommand{\BIBforeignlanguage}[2]{{%
\expandafter\ifx\csname l@#1\endcsname\relax
\typeout{** WARNING: IEEEtran.bst: No hyphenation pattern has been}%
\typeout{** loaded for the language `#1'. Using the pattern for}%
\typeout{** the default language instead.}%
\else
\language=\csname l@#1\endcsname
\fi
#2}}
\providecommand{\BIBdecl}{\relax}
\BIBdecl

\bibitem{7508986}
J.~Dong and H.~Liu, ``Video stabilization for strict real-time applications,'' \emph{IEEE Transactions on Circuits and Systems for Video Technology}, vol.~27, no.~4, pp. 716--724, 2017.

\bibitem{pliego2021quaternion}
J.~Pliego-Jim{\'e}nez, ``Quaternion-based adaptive control for trajectory tracking of quadrotor unmanned aerial vehicles,'' \emph{International Journal of Adaptive Control and Signal Processing}, vol.~35, no.~5, pp. 628--641, 2021.

\bibitem{10256085}
H.~Xu, L.~Wang, W.~Han, Y.~Yang, J.~Li, Y.~Lu, and J.~Li, ``A survey on uav applications in smart city management: Challenges, advances, and opportunities,'' \emph{IEEE Journal of Selected Topics in Applied Earth Observations and Remote Sensing}, vol.~16, pp. 8982--9010, 2023.

\bibitem{10158513}
T.~M. Tran, T.~N. Vu, T.~V. Nguyen, and K.~Nguyen, ``Uit-adrone: A novel drone dataset for traffic anomaly detection,'' \emph{IEEE Journal of Selected Topics in Applied Earth Observations and Remote Sensing}, vol.~16, pp. 5590--5601, 2023.

\bibitem{10119181}
Y.~Zhang, C.~Wu, W.~Guo, T.~Zhang, and W.~Li, ``Cfanet: Efficient detection of uav image based on cross-layer feature aggregation,'' \emph{IEEE Transactions on Geoscience and Remote Sensing}, vol.~61, pp. 1--11, 2023.

\bibitem{9119818}
H.~Chen, T.~Gao, G.~Qian, W.~Chen, and Y.~Zhang, ``Tensored generalized hough transform for object detection in remote sensing images,'' \emph{IEEE Journal of Selected Topics in Applied Earth Observations and Remote Sensing}, vol.~13, pp. 3503--3520, 2020.

\bibitem{dai2024simvg}
M.~Dai, L.~Yang, Y.~Xu, Z.~Feng, and W.~Yang, ``Simvg: A simple framework for visual grounding with decoupled multi-modal fusion,'' \emph{arXiv preprint arXiv:2409.17531}, 2024.

\bibitem{6668877}
M.~E. Angelopoulou and C.-S. Bouganis, ``Vision-based egomotion estimation on fpga for unmanned aerial vehicle navigation,'' \emph{IEEE Transactions on Circuits and Systems for Video Technology}, vol.~24, no.~6, pp. 1070--1083, 2014.

\bibitem{10620851}
J.~M. Kong and E.~Sousa, ``Piggybacking on uav package delivery systems to simultaneously provide wireless coverage: A deep reinforcement learning-based trajectory design,'' in \emph{IEEE INFOCOM 2024 - IEEE Conference on Computer Communications Workshops (INFOCOM WKSHPS)}, 2024, pp. 1--6.

\bibitem{9933782}
Y.~Qin, M.~A. Kishk, and M.-S. Alouini, ``Stochastic-geometry-based analysis of multipurpose uavs for package and data delivery,'' \emph{IEEE Internet of Things Journal}, vol.~10, no.~5, pp. 4664--4676, 2023.

\bibitem{9700861}
L.~Kloeker, T.~Moers, L.~Vater, A.~Zlocki, and L.~Eckstein, ``Utilization and potentials of unmanned aerial vehicles (uavs) in the field of automated driving: A survey,'' in \emph{2021 5th International Conference on Vision, Image and Signal Processing (ICVISP)}, 2021, pp. 9--17.

\bibitem{10194407}
B.~Jiao, L.~Yang, L.~Gao, P.~Wang, S.~Zhang, and Y.~Zhang, ``Vehicle re-identification in aerial images and videos: Dataset and approach,'' \emph{IEEE Transactions on Circuits and Systems for Video Technology}, vol.~34, no.~3, pp. 1586--1603, 2024.

\bibitem{9919263}
Y.~Sun, Z.~Shao, G.~Cheng, X.~Huang, and Z.~Wang, ``Road and car extraction using uav images via efficient dual contextual parsing network,'' \emph{IEEE Transactions on Geoscience and Remote Sensing}, vol.~60, pp. 1--13, 2022.

\bibitem{10582419}
M.~Khurshid, M.~Shahzad, H.~A. Khattak, M.~I. Malik, and M.~M. Fraz, ``Vision-based 3-d localization of uav using deep image matching,'' \emph{IEEE Journal of Selected Topics in Applied Earth Observations and Remote Sensing}, vol.~17, pp. 12\,020--12\,030, 2024.

\bibitem{10177194}
T.~Wang, J.~Li, and C.~Sun, ``Dehi: A decoupled hierarchical architecture for unaligned ground-to-aerial geo-localization,'' \emph{IEEE Transactions on Circuits and Systems for Video Technology}, vol.~34, no.~3, pp. 1927--1940, 2024.

\bibitem{9913952}
X.~Lu, S.~Luo, and Y.~Zhu, ``It’s okay to be wrong: Cross-view geo-localization with step-adaptive iterative refinement,'' \emph{IEEE Transactions on Geoscience and Remote Sensing}, vol.~60, pp. 1--13, 2022.

\bibitem{10601183}
W.-J. Ahn, S.-Y. Park, D.-S. Pae, H.-D. Choi, and M.-T. Lim, ``Bridging viewpoints in cross-view geo-localization with siamese vision transformer,'' \emph{IEEE Transactions on Geoscience and Remote Sensing}, vol.~62, pp. 1--12, 2024.

\bibitem{wang2021each}
T.~Wang, Z.~Zheng, C.~Yan, J.~Zhang, Y.~Sun, B.~Zheng, and Y.~Yang, ``Each part matters: Local patterns facilitate cross-view geo-localization,'' \emph{IEEE Transactions on Circuits and Systems for Video Technology}, vol.~32, no.~2, pp. 867--879, 2021.

\bibitem{10387514}
H.~Zhao, K.~Ren, T.~Yue, C.~Zhang, and S.~Yuan, ``Transfg: A cross-view geo-localization of satellite and uavs imagery pipeline using transformer-based feature aggregation and gradient guidance,'' \emph{IEEE Transactions on Geoscience and Remote Sensing}, vol.~62, pp. 1--12, 2024.

\bibitem{9583266}
X.~Tian, J.~Shao, D.~Ouyang, and H.~T. Shen, ``Uav-satellite view synthesis for cross-view geo-localization,'' \emph{IEEE Transactions on Circuits and Systems for Video Technology}, vol.~32, no.~7, pp. 4804--4815, 2022.

\bibitem{10376356}
M.~Dai, E.~Zheng, Z.~Feng, L.~Qi, J.~Zhuang, and W.~Yang, ``Vision-based uav self-positioning in low-altitude urban environments,'' \emph{IEEE Transactions on Image Processing}, vol.~33, pp. 493--508, 2024.

\bibitem{chen2024sdpl}
Q.~Chen, T.~Wang, Z.~Yang, H.~Li, R.~Lu, Y.~Sun, B.~Zheng, and C.~Yan, ``Sdpl: Shifting-dense partition learning for uav-view geo-localization,'' \emph{arXiv preprint arXiv:2403.04172}, 2024.

\bibitem{zheng2020university}
Z.~Zheng, Y.~Wei, and Y.~Yang, ``University-1652: A multi-view multi-source benchmark for drone-based geo-localization,'' in \emph{Proceedings of the 28th ACM international conference on Multimedia}, 2020, pp. 1395--1403.

\bibitem{10185134}
T.~Shen, Y.~Wei, L.~Kang, S.~Wan, and Y.-H. Yang, ``Mccg: A convnext-based multiple-classifier method for cross-view geo-localization,'' \emph{IEEE Transactions on Circuits and Systems for Video Technology}, vol.~34, no.~3, pp. 1456--1468, 2024.

\bibitem{9779991}
J.~Lin, Z.~Zheng, Z.~Zhong, Z.~Luo, S.~Li, Y.~Yang, and N.~Sebe, ``Joint representation learning and keypoint detection for cross-view geo-localization,'' \emph{IEEE Transactions on Image Processing}, vol.~31, pp. 3780--3792, 2022.

\bibitem{cheng2023ai}
N.~Cheng, S.~Wu, X.~Wang, Z.~Yin, C.~Li, W.~Chen, and F.~Chen, ``Ai for uav-assisted iot applications: A comprehensive review,'' \emph{IEEE Internet of Things Journal}, vol.~10, no.~16, pp. 14\,438--14\,461, 2023.

\bibitem{shadiev2023systematic}
R.~Shadiev and S.~Yi, ``A systematic review of uav applications to education,'' \emph{Interactive Learning Environments}, vol.~31, no.~10, pp. 6165--6194, 2023.

\bibitem{molina2023review}
A.~A. Molina, Y.~Huang, and Y.~Jiang, ``A review of unmanned aerial vehicle applications in construction management: 2016--2021,'' \emph{Standards}, vol.~3, no.~2, pp. 95--109, 2023.

\bibitem{bakirci2024enhancing}
M.~Bakirci, ``Enhancing vehicle detection in intelligent transportation systems via autonomous uav platform and yolov8 integration,'' \emph{Applied Soft Computing}, vol. 164, p. 112015, 2024.

\bibitem{krizhevsky2012imagenet}
A.~Krizhevsky, I.~Sutskever, and G.~E. Hinton, ``Imagenet classification with deep convolutional neural networks,'' \emph{Advances in neural information processing systems}, vol.~25, 2012.

\bibitem{lecun1998gradient}
Y.~LeCun, L.~Bottou, Y.~Bengio, and P.~Haffner, ``Gradient-based learning applied to document recognition,'' \emph{Proceedings of the IEEE}, vol.~86, no.~11, pp. 2278--2324, 1998.

\bibitem{bromley1993signature}
J.~Bromley, I.~Guyon, Y.~LeCun, E.~S{\"a}ckinger, and R.~Shah, ``Signature verification using a" siamese" time delay neural network,'' \emph{Advances in neural information processing systems}, vol.~6, 1993.

\bibitem{hadsell2006dimensionality}
R.~Hadsell, S.~Chopra, and Y.~LeCun, ``Dimensionality reduction by learning an invariant mapping,'' in \emph{2006 IEEE computer society conference on computer vision and pattern recognition (CVPR'06)}, vol.~2.\hskip 1em plus 0.5em minus 0.4em\relax IEEE, 2006, pp. 1735--1742.

\bibitem{hu2018cvm}
S.~Hu, M.~Feng, R.~M. Nguyen, and G.~H. Lee, ``Cvm-net: Cross-view matching network for image-based ground-to-aerial geo-localization,'' in \emph{Proceedings of the IEEE Conference on Computer Vision and Pattern Recognition}, 2018, pp. 7258--7267.

\bibitem{dosovitskiy2020image}
A.~Dosovitskiy, ``An image is worth 16x16 words: Transformers for image recognition at scale,'' \emph{arXiv preprint arXiv:2010.11929}, 2020.

\bibitem{dai2021transformer}
M.~Dai, J.~Hu, J.~Zhuang, and E.~Zheng, ``A transformer-based feature segmentation and region alignment method for uav-view geo-localization,'' \emph{IEEE Transactions on Circuits and Systems for Video Technology}, vol.~32, no.~7, pp. 4376--4389, 2021.

\bibitem{zhu2022transgeo}
S.~Zhu, M.~Shah, and C.~Chen, ``Transgeo: Transformer is all you need for cross-view image geo-localization,'' in \emph{Proceedings of the IEEE/CVF Conference on Computer Vision and Pattern Recognition}, 2022, pp. 1162--1171.

\bibitem{zhang2023cross}
X.~Zhang, X.~Li, W.~Sultani, Y.~Zhou, and S.~Wshah, ``Cross-view geo-localization via learning disentangled geometric layout correspondence,'' in \emph{Proceedings of the AAAI Conference on Artificial Intelligence}, vol.~37, no.~3, 2023, pp. 3480--3488.

\bibitem{deuser2023sample4geo}
F.~Deuser, K.~Habel, and N.~Oswald, ``Sample4geo: Hard negative sampling for cross-view geo-localisation,'' in \emph{Proceedings of the IEEE/CVF International Conference on Computer Vision}, 2023, pp. 16\,847--16\,856.

\bibitem{shi2022cvlnet}
Y.~Shi, X.~Yu, S.~Wang, and H.~Li, ``Cvlnet: Cross-view semantic correspondence learning for video-based camera localization,'' in \emph{Asian Conference on Computer Vision}.\hskip 1em plus 0.5em minus 0.4em\relax Springer, 2022, pp. 123--141.

\bibitem{ye2024sg}
J.~Ye, Q.~Luo, J.~Yu, H.~Zhong, Z.~Zheng, C.~He, and W.~Li, ``Sg-bev: Satellite-guided bev fusion for cross-view semantic segmentation,'' in \emph{Proceedings of the IEEE/CVF Conference on Computer Vision and Pattern Recognition}, 2024, pp. 27\,748--27\,757.

\bibitem{liu2024adaptive}
C.~Liu, S.~Li, C.~Du, and H.~Qiu, ``Adaptive global embedding learning: A two-stage framework for uav-view geo-localization,'' \emph{IEEE Signal Processing Letters}, 2024.

\bibitem{zhuang2021faster}
J.~Zhuang, M.~Dai, X.~Chen, and E.~Zheng, ``A faster and more effective cross-view matching method of uav and satellite images for uav geolocalization,'' \emph{Remote Sensing}, vol.~13, no.~19, p. 3979, 2021.

\bibitem{zhu2023sues}
R.~Zhu, L.~Yin, M.~Yang, F.~Wu, Y.~Yang, and W.~Hu, ``Sues-200: A multi-height multi-scene cross-view image benchmark across drone and satellite,'' \emph{IEEE Transactions on Circuits and Systems for Video Technology}, vol.~33, no.~9, pp. 4825--4839, 2023.

\bibitem{dai2022finding}
M.~Dai, J.~Chen, Y.~Lu, W.~Hao, and E.~Zheng, ``Finding point with image: An end-to-end benchmark for vision-based uav localization,'' \emph{arXiv preprint arXiv:2208.06561}, 2022.

\bibitem{wang2023wamf}
G.~Wang, J.~Chen, M.~Dai, and E.~Zheng, ``Wamf-fpi: A weight-adaptive multi-feature fusion network for uav localization,'' \emph{Remote Sensing}, vol.~15, no.~4, p. 910, 2023.

\bibitem{chen2024fpi}
J.~Chen, E.~Zheng, M.~Dai, Y.~Chen, and Y.~Lu, ``Os-fpi: A coarse-to-fine one-stream network for uav geo-localization,'' \emph{IEEE Journal of Selected Topics in Applied Earth Observations and Remote Sensing}, 2024.

\bibitem{liu2022convnet}
Z.~Liu, H.~Mao, C.-Y. Wu, C.~Feichtenhofer, T.~Darrell, and S.~Xie, ``A convnet for the 2020s,'' in \emph{Proceedings of the IEEE/CVF conference on computer vision and pattern recognition}, 2022, pp. 11\,976--11\,986.

\bibitem{woo2018cbam}
S.~Woo, J.~Park, J.-Y. Lee, and I.~S. Kweon, ``Cbam: Convolutional block attention module,'' in \emph{Proceedings of the European conference on computer vision (ECCV)}, 2018, pp. 3--19.

\bibitem{schroff2015facenet}
F.~Schroff, D.~Kalenichenko, and J.~Philbin, ``Facenet: A unified embedding for face recognition and clustering,'' in \emph{Proceedings of the IEEE conference on computer vision and pattern recognition}, 2015, pp. 815--823.

\bibitem{ding2020practical}
L.~Ding, J.~Zhou, L.~Meng, and Z.~Long, ``A practical cross-view image matching method between uav and satellite for uav-based geo-localization,'' \emph{Remote Sensing}, vol.~13, no.~1, p.~47, 2020.

\bibitem{10835173}
J.~Yang, D.~Qin, H.~Tang, S.~Tao, H.~Bie, and L.~Ma, ``Dinov2-based uav visual self-localization in low-altitude urban environments,'' \emph{IEEE Robotics and Automation Letters}, vol.~10, no.~2, pp. 2080--2087, 2025.

\bibitem{10506984}
C.~Liu, S.~Li, C.~Du, and H.~Qiu, ``Adaptive global embedding learning: A two-stage framework for uav-view geo-localization,'' \emph{IEEE Signal Processing Letters}, vol.~31, pp. 1239--1243, 2024.

\bibitem{liu2021swin}
Z.~Liu, Y.~Lin, Y.~Cao, H.~Hu, Y.~Wei, Z.~Zhang, S.~Lin, and B.~Guo, ``Swin transformer: Hierarchical vision transformer using shifted windows,'' in \emph{Proceedings of the IEEE/CVF international conference on computer vision}, 2021, pp. 10\,012--10\,022.

\bibitem{liu2022swin}
Z.~Liu, H.~Hu, Y.~Lin, Z.~Yao, Z.~Xie, Y.~Wei, J.~Ning, Y.~Cao, Z.~Zhang, L.~Dong \emph{et~al.}, ``Swin transformer v2: Scaling up capacity and resolution,'' in \emph{Proceedings of the IEEE/CVF conference on computer vision and pattern recognition}, 2022, pp. 12\,009--12\,019.

\bibitem{10129939}
B.~Sun, G.~Liu, and Y.~Yuan, ``F3-net: Multiview scene matching for drone-based geo-localization,'' \emph{IEEE Transactions on Geoscience and Remote Sensing}, vol.~61, pp. 1--11, 2023.

\bibitem{10288351}
H.~Song, Z.~Wang, Y.~Lei, D.~Shi, X.~Tong, Y.~Lei, and C.~Qiu, ``Learning visual representation clusters for cross-view geo-location,'' \emph{IEEE Geoscience and Remote Sensing Letters}, vol.~20, pp. 1--5, 2023.

\bibitem{10517694}
F.~Ge, Y.~Zhang, L.~Wang, W.~Liu, Y.~Liu, S.~Coleman, and D.~Kerr, ``Multilevel feedback joint representation learning network based on adaptive area elimination for cross-view geo-localization,'' \emph{IEEE Transactions on Geoscience and Remote Sensing}, vol.~62, pp. 1--15, 2024.

\bibitem{10506965}
Q.~Li, X.~Yang, J.~Fan, R.~Lu, B.~Tang, S.~Wang, and S.~Su, ``Geoformer: An effective transformer-based siamese network for uav geolocalization,'' \emph{IEEE Journal of Selected Topics in Applied Earth Observations and Remote Sensing}, vol.~17, pp. 9470--9491, 2024.

\end{thebibliography}

\vfill

\end{document}